%% file: egpaper.tex
\documentclass[10pt,twocolumn,letterpaper]{article}

\usepackage{wacv}
\usepackage{times}
\usepackage{epsfig}
\usepackage{graphicx}
\usepackage{amsmath}
\usepackage{amssymb}
\usepackage[mathscr]{euscript}
\usepackage{booktabs}
\usepackage{authblk}
\usepackage[hyphens]{url}
\newcommand{\new}[1]{\textcolor{black}{#1}} %

\def\wacvPaperID{2091} %

\wacvapplicationstrack %

\wacvfinalcopy %

\ifwacvfinal
\usepackage[breaklinks=true,bookmarks=false]{hyperref}
\else
\usepackage[pagebackref=true,breaklinks=true,colorlinks,bookmarks=false]{hyperref}
\fi

\begin{document}

\title{Watch Those Words: \\Video Falsification Detection Using Word-Conditioned Facial Motion}

\author[1]{Shruti Agarwal\thanks{shrutia@mit.edu}}
\author[2]{Liwen Hu}
\author[1]{Evonne Ng}
\author[1]{Trevor Darrell}
\author[2]{Hao Li}
\author[1]{Anna Rohrbach}
\affil[1]{University of California, Berkeley}
\affil[2]{Pinscreen, Inc.}

\maketitle
\thispagestyle{empty}

\begin{abstract}
   In today's era of digital misinformation, we are increasingly faced with new threats posed by video falsification techniques. Such falsifications range from cheapfakes (e.g., lookalikes or audio dubbing) to deepfakes (e.g., sophisticated AI media synthesis methods), which are becoming perceptually indistinguishable from real videos. To tackle this challenge, we propose a multi-modal semantic forensic approach to discover clues that go beyond detecting discrepancies in visual quality, thereby handling both simpler cheapfakes and visually persuasive deepfakes. In this work, our goal is to verify that the purported person seen in the video is indeed themselves by detecting anomalous facial movements corresponding to the spoken words. We leverage the idea of attribution to learn person-specific biometric patterns that distinguish a given speaker from others. We use interpretable Action Units (AUs) to capture a person's face and head movement as opposed to deep CNN features, and we are the first to use word-conditioned facial motion analysis. We further demonstrate our method's effectiveness on a range of fakes not seen in training including those without video manipulation, that were not addressed in prior work.
\end{abstract}

\maketitle

\begin{figure}
    \begin{center}
        \begin{tabular}{c@{\hspace{0.1cm}}c@{\hspace{0.1cm}}c}
        \multicolumn{3}{c}{Video of Obama saying ``Hi'' (as in ``Hi Everybody'')} \\
        \includegraphics[width=0.15\textwidth]{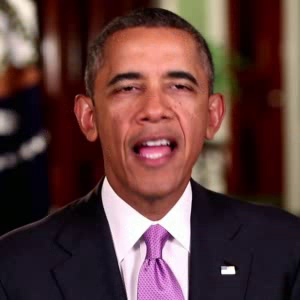} &
        \includegraphics[width=0.15\textwidth]{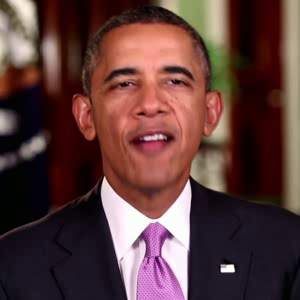} &
        \includegraphics[width=0.15\textwidth]{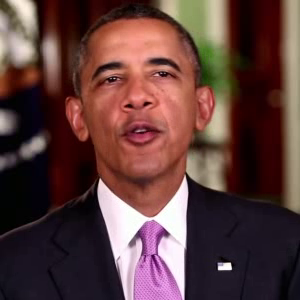} \\
        ``Hi''  & [h] & [a\textsc{i}] \\
        \includegraphics[width=0.15\textwidth]{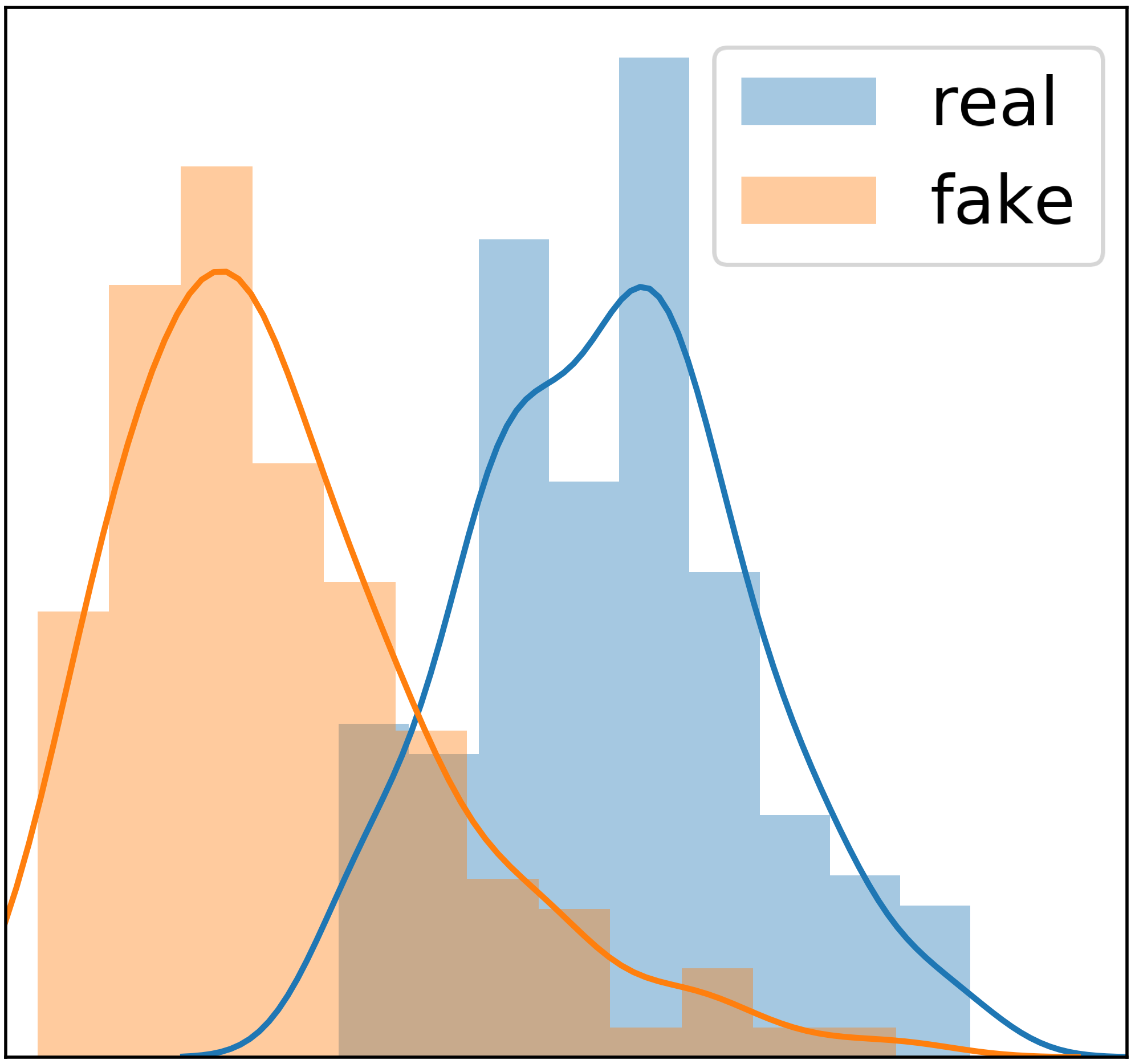} &
        \includegraphics[width=0.15\textwidth]{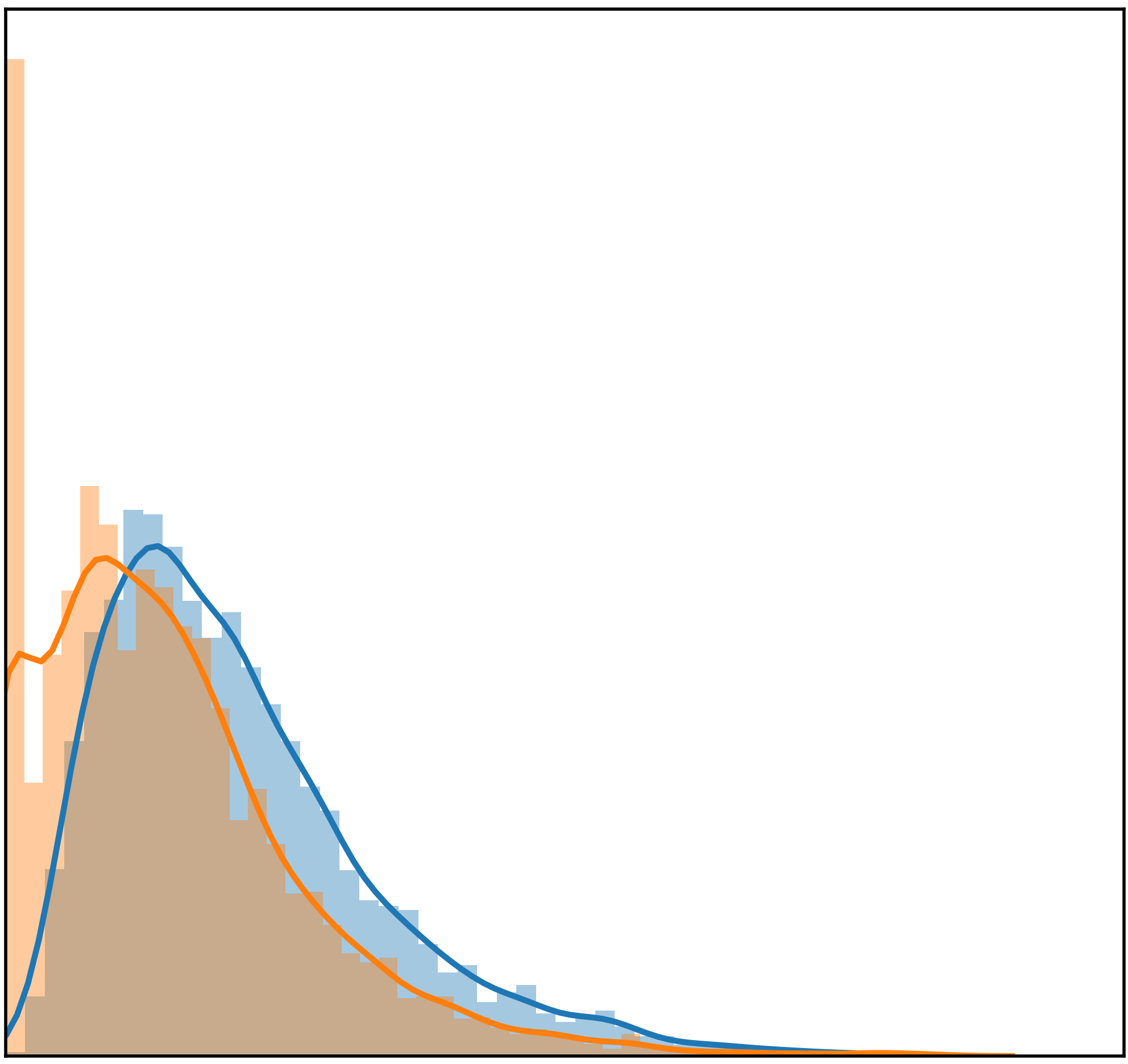} &
        \includegraphics[width=0.15\textwidth]{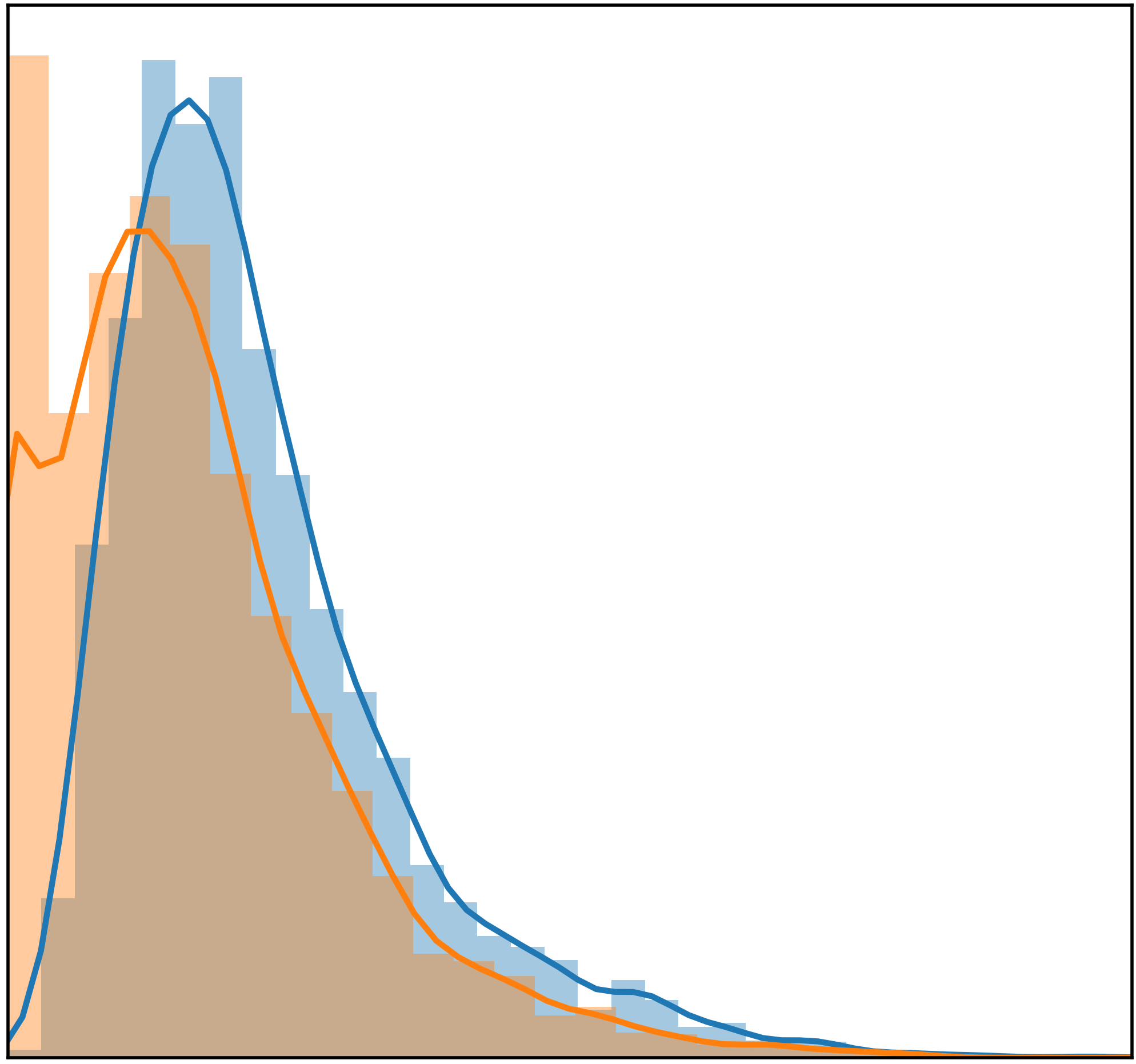} \\
        \multicolumn{3}{c}{head rotation along X-axis}
        \end{tabular}
    \end{center}
    \vspace{-0.3cm}
    \caption{\new{We are looking for inconsistencies between \emph{what they say} and \emph{how they move}. One way to address this is via audio, or sounds (phonemes), however this may miss some important semantic cues. Consider an example of Obama saying ``Hi'' (as in ``Hi Everybody''). We noticed that each time he says that, he rotates his head along X-axis. If we plot the amount of ``head rotation along X-axis'' for each occurrence of the word ``Hi'', we see a clear separation between the real and fake Obama videos, allowing us to successfully detect falsification. At the same time, the two phonemes, [h] and [a\textsc{i}], that constitute the word ``Hi'', do not have any correlation with head rotation. This is intuitive since these phonemes also occur in many other words, where Obama does not move his head. This motivates us to focus on spoken \emph{words} to discover biometric word-specific patterns.}}
    \vspace{-0.6cm}
    \label{fig:teaser}
\end{figure}
\input{intro}

\input{related}

\input{methods}
\input{dataset}

\input{results}
\input{conclusion}

{\small
\bibliographystyle{ieee_fullname}
\bibliography{egbib}
}

\clearpage
\appendix
\input{appendix_arxiv}

\end{document}

%% file: intro.tex
\section{Introduction}
\label{sec:intro}

Humans tend to trust what they see, especially when it comes to video. Historically, video has been the best proof that an event has indeed occurred. However, in the rapidly evolving misinformation landscape of the present digital era, this may not be true for long. Video manipulation techniques are more accessible than ever, while the reach of internet and social media enables rapid spread of falsified content. Recent headlines, such as \emph{"XR Belgium posts deepfake of Belgian premier linking Covid-19 with climate crisis"}~\cite{xrbelgium}, \emph{``Dutch MPs in video conference with deepfake imitation of Navalny's Chief of Staff''}~\cite{nltimes}\footnote{A later report explained that in fact it was an impersonator ~\cite{theverge}.}, \emph{``When virtual turns fake: Danish politicians 'meet' Belarusian opposition figure''}~\cite{reuters} are examples of how both deepfakes and cheapfakes (e.g. lookalikes) pose real threats and have serious consequences, especially if targeted at people in power.

To protect the public against potential disinformation campaigns, new deepfake detection methods are being introduced to combat new and more advanced deepfake techniques~\cite{suwajanakorn2017,Koki2018,kim-etal-siggraph-18,nirkin2019fsgan,fried2019text,perov2020deepfacelab,prajwal2020lip,thies2020neural}.
Not only is detection increasingly challenging, most methods are ineffective against cheapfakes falsified through conventional techniques (e.g., speeding up or slowing down a video), or with no video manipulation at all (a lookalike, audio dubbing).

In this work, we aim to detect video falsifications related to a person's identity. Specifically, we aim to detect if the purported person ``seen'' in a video is indeed themselves. This is distinct from the deepfake detection problem, where the goal is to distinguish between pristine (non-manipulated) and generated/altered videos. For such methods, a video of an impersonator would be wrongly identified as ``real''. Similarly, a non-manipulated video with dubbed or edited speech would also be considered ``real'' by many deepfake detectors, since often they do not take speech into account. In contrast, our problem statement is more general, as it includes both deepfakes and falsified pristine videos. Furthermore, as the deepfakes' quality improves, detecting visual flaws will also become increasingly difficult. 
Our key insight is to use \emph{semantic, person-specific cues as an alternative, and generalizable solution to detecting video falsifications.} %

Biometric-based techniques~\cite{agarwal2019} have been recently introduced to identify falsified videos, that are either not manipulated or extremely realistic.~\cite{agarwal2019} analyze the authenticity of a person by correlating the head and facial movements in the existing footage of the person. Even though several hours of training video is required, 
they are well-suited for public figures such as celebrities and world leaders who are often a target. 
However, these person-specific methods are ineffective against cutting-edge audio-to-lip synthesis techniques such as~\cite{suwajanakorn2017,prajwal2020lip} or commercial video dialog replacement solutions developed by Synthesia\footnote{\href{https://www.synthesia.io/}{https://www.synthesia.io/}} or Canny AI\footnote{\href{https://www.cannyai.com/}{https://www.cannyai.com/}}, which only manipulate the mouth.
To this end, we propose a semantic, multimodal detection approach that integrates speech transcripts into person-specific gesture analysis. We leverage interpretable Action Units (AUs)~\cite{baltruvsaitis2015cross} to model a person's face and head movement. Our approach is to analyze the \emph{word-conditioned facial movement} captured with AUs to learn per-word models for classifying real and fake videos. %
Our intuition is that each individual may have identifying, %
unique patterns in how their speech, facial expressions, and gestures co-occur. %
\new{This is distinct from using raw audio or sounds, as illustrated in Figure~\ref{fig:teaser}. The individual phonemes lack the semantics and thus may not capture high-level idiosyncratic regularities of facial gestures associated with specific words}. %
Our word-level models distinguish real and falsified videos using the facial patterns associated with specific words. At test time, we compute classification scores for each word in a video clip, and aggregate them into a final score.

We experiment using real/fake videos of world leaders and TV talk-show hosts, where we consider a full spectrum of cutting edge video manipulation techniques~\cite{prajwal2020lip,perov2020deepfacelab,suwajanakorn2017}, as well as fakes found in the wild.
We compare our approach to several prominent prior works, and we show that we achieve the best performance across the entire range of fakes. No other method that we consider demonstrates such general capability, as they tend to suffer on audio dubbing or in-the-wild lip-sync fakes.\new{We additionally compare using phonemes instead of words within our model design, and see that while it performs well on cases with audio-visual inconsistencies, it struggles to recognize the fakes that require biometric features.} We perform ablation studies to confirm that the key advantage of our method is indeed the word-conditioned analysis. 
Lastly, an added benefit of our approach is interpretability: we are able to capture human understandable, person-specific word-movement patterns  predictive of a video being real or falsified (e.g., common in real videos but absent in the fakes ones).

Our contributions are as follows. (a) %
We present a new, general problem statement: given a video, predict whether a person is authentic, regardless if falsification is a deep- or cheapfake. (b) We propose the first semantic person-specific approach to address this problem that leverages word-conditioned facial movements. %
(c) We perform a comparative study across multiple fake types, ranging from deepfakes to impersonators and audio dubbing. Unlike prior work, our approach shows strong generalization across all types of fakes. \new{Namely, our method exhibits two key capabilities: recognizing speech-video inconsistency, while also capturing biometric features.} (d) Our approach also offers interpretability, allowing us to expose the person-specific predictive word-gesture patterns.

%% file: related.tex
\section{Related Work}
\label{sec:related}

We identify two types of detection techniques: (1) Person-generic methods analyze whether manipulation occurred regardless of the person's identity; (2) Person-specific methods verify whether the characteristics of the seen individual match the real person. Person-generic approaches are often trained on large datasets with real and fake videos, and rely on either \textbf{low-level features} or \textbf{high-level semantics}. Person-specific methods, on the other hand, typically require additional \textbf{biometric-based} data for identification.

\noindent\textbf{Low-Level Feature-Based Forensics.}
These methods (often CNN classifiers) are typically person-generic and focus on visual artifacts or statistical anomalies learned implicitly from images or videos~\cite{afchar2018,8639163,nguyen2019capsule,yu2019attributing,8638330,rossler2019faceforensics++,zhao2021multi, zhou17twostream, nguyen2019capsule, qian2020thinking, wang2021m2tr}. 
While many techniques struggle to generalize to new video manipulation techniques or unseen deepfake videos~\cite{dolhansky2019deepfake}, some focus on artifacts that appear also for unseen fakes.~\cite{li2018warping} detect the warping artifacts, ~\cite{li2019xray} identify blending traces during synthesis, and ~\cite{wang20easy} leverage inconsistencies between images and meta-data. While promising detection capabilities have been shown, these methods are often susceptible to deteriorations like compression, resolution reduction, or adversarial perturbations and attacks~\cite{carlini2020evading,Hussain2021}. 

\noindent\textbf{High-Level Semantic-Based Forensics.}
Person-generic and high-level semantic-based techniques focus on explicit anomalies of person's characteristic, such as the absence of eye blinking~\cite{li2018blinking}, the inconsistencies in head pose~\cite{yang19headpose}, human physiological signals like heart beat~\cite{fernandes2019predicting, qi2020deeprhythm}, ear movements~\cite{agarwal2021aural}, and other biological signals~\cite{ciftci2020fakecatcher}. These approaches often generalize better to unseen deepfakes and are more resilient to laundering. However, a reliable extraction of high-level features is often difficult to achieve in unconstrained settings and short video clips.
Several recent methods focus on temporal inconsistencies in facial performances~\cite{haliassos2021lips, chugh2020not, mittal2020emotions, agarwal2020viseme} but rely on robust 3D face tracking. 

Similar to our proposed work, multi-modal techniques~\cite{chugh2020not, mittal2020emotions, zhou2021joint} exploit both the audio and visual signal to detect deepfakes. While audio signals can provide cues like emotions and \emph{how} a person is talking, our work focuses on spoken words, which provide a more direct information about \emph{what} is being said. %
For example, different head nods associated with words convey agreement, disagreement or greetings in many cultures~\cite{darwin2015expression}. We believe that our approach is complementary to audio-based methods, since it captures word-specific patterns, inaccessible to raw audio, as illustrated in Figure~\ref{fig:teaser}.
In~\cite{agarwal2020viseme}, the authors exploit the shape of the lips when phonemes 'P', 'B', or 'M' are being pronounced. Whereas in~\cite{haliassos2021lips}, the authors use only visual signal to detect if the lip movements are `readable'' in a video. Even though these techniques can detect deepfakes where the lips are modified, since they are not person-specific they will struggle identifying video falsifications using impersonators.%

\noindent\textbf{Biometric-Based Forensics.}
Biometric-based detection methods~\cite{1167045,5595938,agarwal2019, cozzolino2020id, agarwal2020detecting, yang2020preventing, korshunov2018speaker} are person-specific as they try to verify the authenticity of a person using known identity priors. These works are the most relevant to our technique and many of them exploit person-specific facial movement over time to detect deepfakes. In~\cite{korshunov2018speaker, yang2020preventing}, the authors use visual appearance and movement of lips to perform speaker verification and detect person-specific deepfakes. In these previous works, the authors analysed only a small set of words which will restrict their approach to fakes where those are being said. In contrast, we include the facial movements from the entire face and use a much larger vocabulary size, enabling our approach to handle in-the-wild deepfakes. The method of ~\cite{agarwal2019} introduced a biometric approach for public figures, where person-specific facial movements in a video are compared with those of pristine videos. Despite the requirement for hours of training data of a known person, this approach is resistant to realistic deepfakes, or even to lookalikes, when no video manipulation is used. More advanced techniques that incorporate CNN-based behavior classification using optical flow~\cite{agarwal2020detecting} or 3DMM-based facial tracking~\cite{cozzolino2020id} have shown improved performance for deepfake detection. Nevertheless, recent advancements in speech-to-lip synthesis~\cite{suwajanakorn2017,prajwal2020lip} show that it is possible to produce highly convincing speech manipulations without altering global facial characteristics. %
In this work, we introduce a multi-modal semantic approach that exploits the fact that spoken words may be associated with distinct person-specific facial movements. In particular, these movements involve the entire face/head and not only the lip region of a person, and are difficult to disguise even for skilled impersonators.

%% file: methods.tex
\section{Word-Conditioned Facial Analysis}
\label{sec:method}

Given an input video of an individual, our goal is to classify it as real or fake. %
We leverage the key insight that individuals often use identifying gestures associated with specific interactions like greeting, disagreement, etc. In our approach, we represent these conversational units in terms of words and analyze the facial gestures associated with them. Considering conversational units at the granularity of words gives us a good trade-off between the number of occurrences of each unique unit and speech semantics. Using N-grams or unique sentences would result in fewer occurrences while phonemes would result in less meaningful speech semantics. \new{We include an empirical comparison of our approach to its phoneme-conditioned counterpart in Section~\ref{sec:results}.}

\begin{figure}[t!]
 \centering
 \includegraphics[width=3.25in]{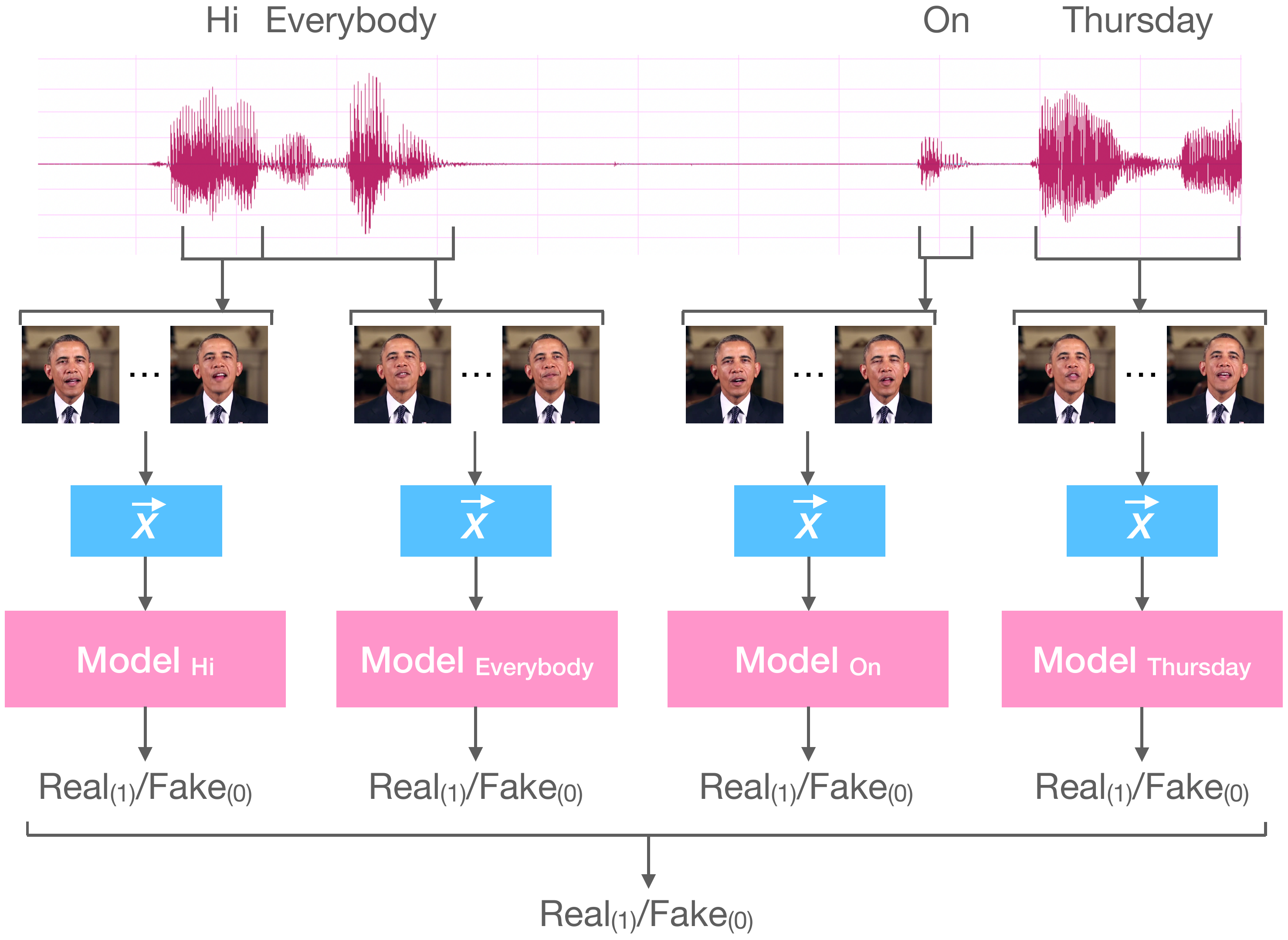}
 \vspace{-0.3cm}
 \caption{Given an input video, we first transcribe the audio and obtain per-frame alignments of word. For each word, an AU-based feature vector $\vec{x}$ is extracted from the corresponding frames $\mathbb{F}$ and then evaluated by a word-specific classifier $\vec{\theta}$. A final score for the possibility of ``Real'' is computed for the video using the geometric mean of all the scores.}
 \label{fig:pipeline}
 \vspace{-0.6cm}
\end{figure}

As shown in Figure~\ref{fig:pipeline}, we first transcribe the audio and then extract the corresponding per-frame facial movements of the speaker represented by AUs~\cite{ekman1976measuring}. 
We encode the speaker motion as the amount of change in the AUs that happens within the window of the word's occurrence. 
Finally, word-level classifiers are trained to detect whether the visual movement match with the spoken words.

\noindent\textbf{Word-Aligned Facial Feature Extraction.}
We denote $\mathbf{F}_{1:T} = \{f\}_{i=1}^{T}$ as a set of frames $f$ from a video of length $T$. Given a video,
we transcribe the audio of the video to get the phonation time of each word $w$, expressed as the start $f_s$ and end $f_n$ frames, where $d=n-s$ is the duration of the phonation.
To associate an individual's facial expression and head motion with the corresponding word, we extract AUs for the window $\mathbf{F}_{s:n}$. %
In contrast to 3-D or 2-D facial landmarks, these AUs represent semantically meaningful micro-expressions such as the strength of a cheek or chin movement (e.g. ``chin raiser''). For a given word spoken within the frame range $\mathbf{F}_{s:n}$, 
we extract a $25$-D facial feature $\vec{g}_i$ at each timestep to obtain $\mathbb{G}_{s-t:n+t}= \{\vec{g}\}_{i=s-t}^{n+t}$. A padding of t=3 frames is added to account for small misalignment between the words and the frames. However, we assume that there is no large misalignment between the video and audio signal.
Each $25$-D facial feature is composed of 4 components: (1) the intensities of the $17$ AUs, (2) the 3-D head rotations and 3-D head translations along X, Y, Z axis, (3) the 3-D horizontal distance between mouth corners (lip-hor), and (4) the 3-D vertical distance between the lower and upper lip (lip-ver).
Instead of using a variable-length feature $\mathbb{G}_{s-t:n+t} \in \mathbb{R}^{d \times 25}$, we use the deltas between the maximum and the minimum values extracted during the word phonation window. 
The facial feature of each word occurrence is then expressed as:
\begin{equation}
    \vec{x}_w = \max_{\vec{g}_i \in \mathbb{G}_{s-t:n+t}}(\vec{g}_i) - \min_{\vec{g}_i \in \mathbb{G}_{s-t:n+t}}(\vec{g}_i),
\label{eq:feature}
\end{equation}
where $\vec{x}_w \in \mathbb{R}^{25\times1}$ is used for building a word-specific model. 
Intuitively, these features capture the maximum range of movement happening when a word is spoken (how much the head moves up when saying ``Hi'') \new{ regardless of the temporal misalignment of features. E.g., in the real videos of Obama the word ``Hi'' spans a minimum of 9 frames and a maximum of 27 frames. By using the range of motion as the feature, we are thus avoiding temporal variability across different utterances of the same word.}

\noindent\textbf{Word-Specific Classifiers.}
We train linear per-word classifiers to tell if the given gesture features belong to the given words. Instead of using more complex learning-based approaches with high-dimensional features, we use linear classifiers to highlight the efficacy of our interpretable features in a simple model.
Given real videos where the words-specific facial movements are correct, we create simulated fakes where words are deliberately matched with random facial movements (speech transcript matched to a wrong video). 
In addition to that, we create synthesized fakes using the recent lip-sync generation method Wav2Lip~\cite{prajwal2020lip}. As before, words are deliberately matched with random facial movements but now the lips are synthesized to say the words. By using these synthesized fakes, we ensure that our classifier does not rely only on lip reading errors in order to detect fakes.
Using this real and fake data, we train person-specific word-specific logistic regression classifiers.

Let $\vec{x}_w \in \mathbb{R}^{25\times1}$ be the facial feature corresponding to word $w$. Let $y_w \in [0,1]$ be the ground truth label of $\vec{x}_w$ where $y_w=1$ if $\vec{x}_w$ is from a real video sequence. %
We learn the model parameters $\boldsymbol{\theta}_w \in \mathbb{R}^{25\times1}$ for a linear classifier that maximizes the following objective function $\mathscr{L}_{\boldsymbol{\theta}_w}$:

\vspace{-3mm}
\begin{equation}
    \mathscr{L}_{\boldsymbol{\theta}_w} = \prod_{i=1}^{M} P(y_i|\vec{x}_i), 
\end{equation}
where $P(y_i|\vec{x}_i)$ is the probability of $y_i$ given $\vec{x}_i$ and $M$ is the number of total occurrences of $w$ in the training data. 

\begin{equation}
    P(y_i|\vec{x}_i) = [\sigma(\boldsymbol{\theta}_w^\top \cdot \vec{x}_i)]^{y_i}.[1-\sigma(\boldsymbol{\theta}_w^\top \cdot \vec{x}_i)]^{1-y_i}
\end{equation}
where $\sigma(x) = \frac{1}{1+e^{-x}}$ is the sigmoid function. 

\noindent\textbf{Testing.}
During evaluation, given a %
test video of a purported individual, we extract the features as described above. The transcribed words not seen in the training set are discarded. %
For each remaining word $w$, the corresponding facial features $\vec{x}_w$ are examined using the target word classifier $\boldsymbol{\theta}_w$ for the given individual. A score $s_w$ in the range of 0 (fake) and 1 (real) for $\vec{x}_w$ is computed as:

\vspace{-3mm}
\begin{equation}
    s_w = \sigma(\boldsymbol{\theta}_w^\top \cdot \vec{x}_w).
\end{equation}
A final score is computed for a given video using the geometric mean of scores across all trained words in the video. 

%% file: dataset.tex
\begin{figure}[t!]
    \begin{center}
        \begin{tabular}{c@{\hspace{0.1cm}}c@{\hspace{0.1cm}}c@{\hspace{0.1cm}}c@{\hspace{0.1cm}}c}
        & Real & Wav2Lip & Impersonator & FaceSwap \\
        \rotatebox[origin=lB]{90}{Obama} & 
        \includegraphics[width=0.2\linewidth]{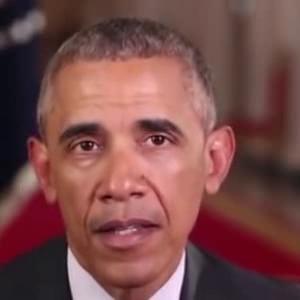} & 
        \includegraphics[width=0.2\linewidth]{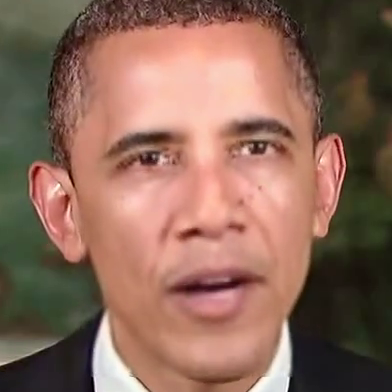} & 
        \includegraphics[width=0.2\linewidth]{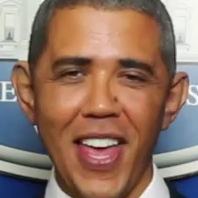} & 
        \includegraphics[width=0.2\linewidth]{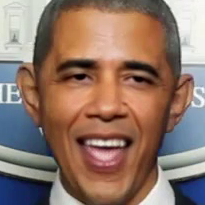}  \\
        
        \rotatebox[origin=lB]{90}{Trump} & 
        \includegraphics[width=0.2\linewidth]{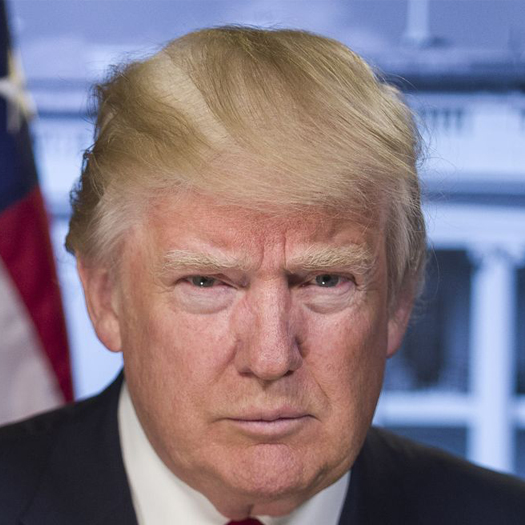} & 
        \includegraphics[width=0.2\linewidth]{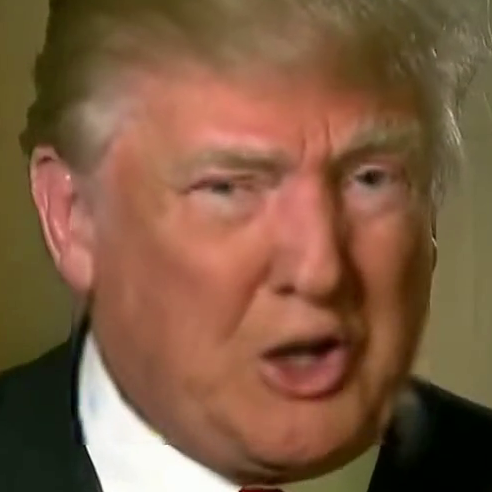} & 
        \includegraphics[width=0.2\linewidth]{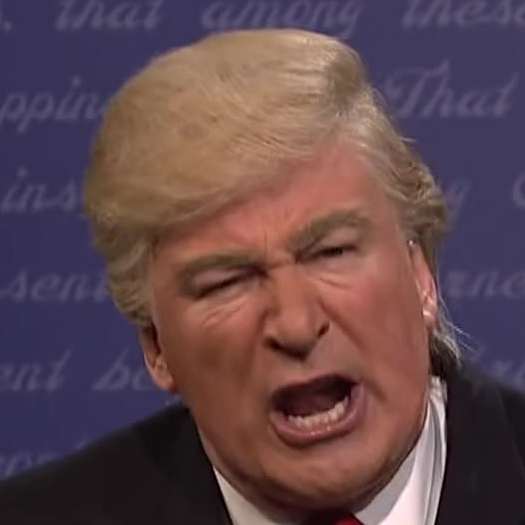} & 
        \includegraphics[width=0.2\linewidth]{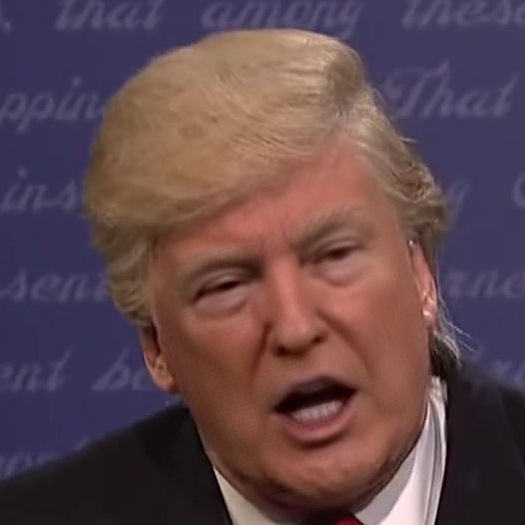}  \\
        
        \rotatebox[origin=lB]{90}{Biden} & 
        \includegraphics[width=0.2\linewidth]{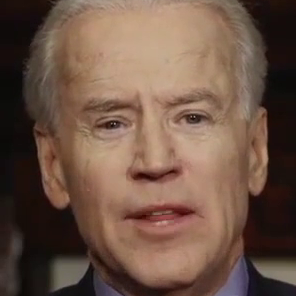} & 
        \includegraphics[width=0.2\linewidth]{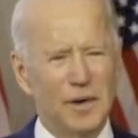} & 
        \includegraphics[width=0.2\linewidth]{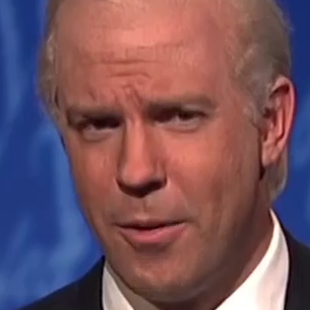} & 
        \includegraphics[width=0.2\linewidth]{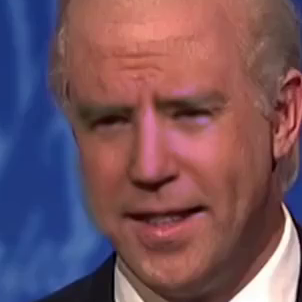}  \\
        
        \rotatebox[origin=lB]{90}{Harris} & 
        \includegraphics[width=0.2\linewidth]{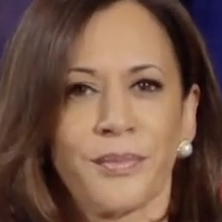} & 
        \includegraphics[width=0.2\linewidth]{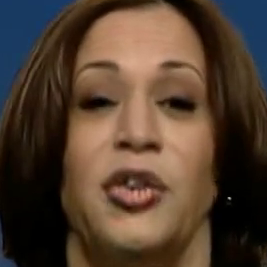} & 
        \includegraphics[width=0.2\linewidth]{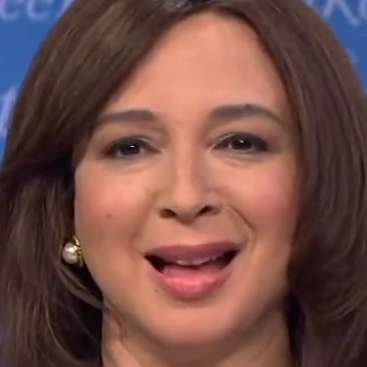} & 
        \includegraphics[width=0.2\linewidth]{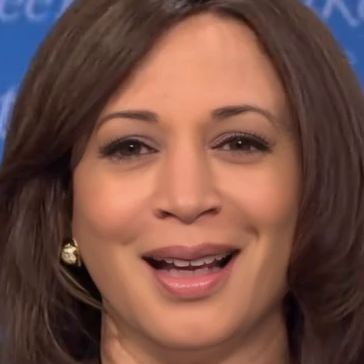}
         \\
        
        \rotatebox[origin=lB]{90}{Oliver} & 
        \includegraphics[width=0.2\linewidth]{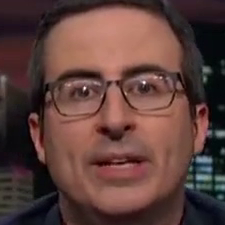} & 
        \includegraphics[width=0.2\linewidth]{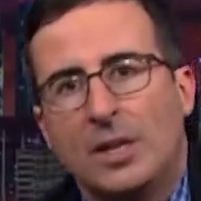} & 
        \includegraphics[width=0.2\linewidth]{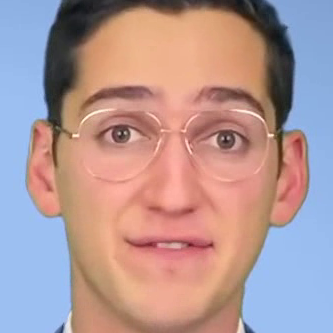} & 
        \includegraphics[width=0.2\linewidth]{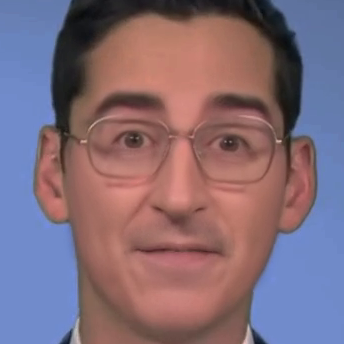} 
         \\
        
        \rotatebox[origin=lB]{90}{O'Brien} & 
        \includegraphics[width=0.2\linewidth]{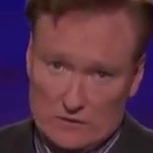} & 
        \includegraphics[width=0.2\linewidth]{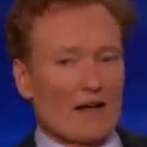} & 
        \includegraphics[width=0.2\linewidth]{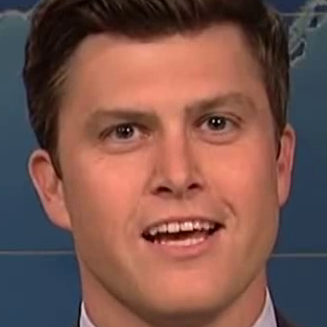} & 
        \includegraphics[width=0.2\linewidth]{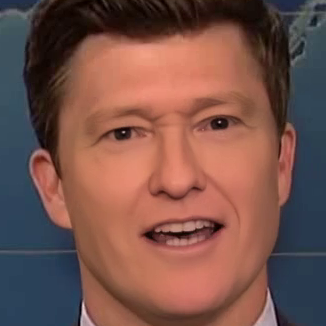} 
         \\
        \end{tabular}
    \end{center}
    \vspace{-0.4cm}
    \caption{Examples of the data used in our work, spanning different types of falsified video, from deepfakes to fakes with non-manipulated video.}
    \label{fig:data_example}
\end{figure}

\begin{figure}
    \begin{center}
        \begin{tabular}{c@{\hspace{0.1cm}}c@{\hspace{0.1cm}}c@{\hspace{0.1cm}}c}
        & Obama & Trump & Biden \\
        \rotatebox[origin=lB]{90}{in-the-wild} & \includegraphics[width=0.2\linewidth]{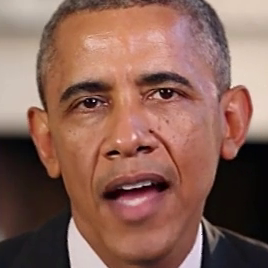}
        & \includegraphics[width=0.2\linewidth]{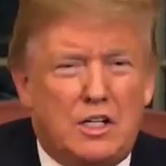}
        & \includegraphics[width=0.2\linewidth]{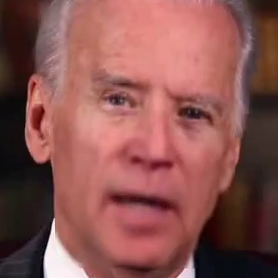} \\
        \end{tabular}
    \end{center}
    \vspace{-0.4cm}
    \caption{Examples of the data used in our work, specifically, in-the-wild lip-sync examples for three individuals.}
\label{fig:itw_example}
\vspace{-0.6cm}
\end{figure}

\section{Dataset}
\label{sec:dataset}

To validate our proposed approach on the general problem statement that includes both deepfakes and non-manipulated fakes, we compile the following dataset. 
We consider four US politicians (Barack Obama, Donald Trump, Joe Biden, Kamala Harris) and two TV talk-show hosts (John Oliver, Conan O' Brien). Further we provide the details for the types of data that we use.

    \noindent\textbf{Real:} %
    The real videos of the politicians were taken from the World Leaders Dataset (WLDR) \cite{agarwal2020detecting} and the videos of the talk-show hosts were taken from~\cite{ginosar2019learning}. The total hours and example frames are shown in Table~\ref{tab:dataset} (Column 1) and Figure~\ref{fig:data_example} (Column 1).
    
    \noindent\textbf{Dubbing:} Using the real videos for each individual, we simulate the dubbing scenario by mismatching the video and the audio. For every real video, a new dubbed video is created by matching it with a random audio of the same length. We produce the same number of hours for the dubbed videos as we have of real videos, Table~\ref{tab:dataset}. 

    \noindent\textbf{Wav2Lip:} Using the real videos, we create lip-sync deepfakes where the lip region in the video is modified to match a random audio. We use the off-the-shelf implementation of Wav2Lip~\cite{prajwal2020lip} to create these fakes. The example frames are shown in Figure~\ref{fig:data_example} (Column 2). 
    
    \begin{table}[t]
    \centering
    \resizebox{0.48\textwidth}{!}
	{
    \begin{tabular}{l|r|c|c|c|c|c}
         & Real & Dubbing & Wav2Lip & Impersonator & FaceSwap & itw \\
         \hline
         Obama &    12.5    &12.5    &12.5    & 0.16 & 0.11 & 0.99 \\
         Trump &    6.1     &6.1     &6.1     & 0.19 & 0.19 & 0.08 \\
         Biden &    5.1     &5.1     &5.1     & 0.04 & 0.14 & 0.12 \\
         Harris &   2.5     &2.5     &2.5     & 0.05 & 0.05 & - \\
         O' Brien & 14.5    &14.5    &14.5    & 0.08 & 0.08 & - \\
         Oliver &   17.8    &17.8    &17.8    & 0.04 & 0.04 & - \\
    \end{tabular}
    }
    \vspace{0.2cm}
    \caption{Number of hours of video for each of the six individuals for different types of video falsification scenarios.}
    \label{tab:dataset}
\end{table}
    
    \noindent\textbf{Impersonator:} The person-specific impersonator videos are obtained from Saturday Night Live on YouTube. The impersonator videos for Obama, Biden, and Trump are from WLDR, and for Harris, Oliver and O'Brien from YouTube. The total hours and example frames are shown in Table~\ref{tab:dataset} (Column 4) and Figure~\ref{fig:data_example} (Column 3).
    
    \noindent\textbf{FaceSwap:} The FaceSwap deepfakes are created using the impersonator videos by replacing impersonator's face with the target person's face. The videos for Obama, Biden, and Trump are from WLDR and for Harris, Oliver and O'Brien are created using the FaceSwap library~\cite{perov2020deepfacelab}. The total hours and example frames are shown in Table~\ref{tab:dataset} (Column 5) and Figure~\ref{fig:data_example} (Column 4).
    
    \noindent\textbf{In-the-wild (itw):} The in-the-wild lip-sync videos for Obama, Trump, and Biden are collected from \cite{agarwal2021aural,agarwal2019,suwajanakorn2017}. The total hours and example frames are shown in Table~\ref{tab:dataset} (Column 6) and Figure~\ref{fig:itw_example}.

%% file: results.tex
\section{Experiments}
\label{sec:result}

\begin{table}[t]
    \centering
    \resizebox{0.48\textwidth}{!}
	{
    \begin{tabular}{l|c|c||c|c|c|c}
    
    & \multicolumn{2}{c||}{Training} & \multicolumn{4}{|c}{Number of unique words during tesing} \\
    \hline
        & Number & Number & Real/ &  &  &  \\
        & of & of & Dubbing/ &  &  &  \\
         & Words & models & Wav2Lip & Impersonator & FaceSwap & itw \\
         \hline
         Obama &   4,925  & 918 & 812 & 248 & 211 & 543  \\
         Trump &   3,664  & 817 & 543 & 296 & 282 & 81 \\
         Biden &   3,985  & 816 & 523 & 133 & 145 & 121 \\
         Harris &  2,270  & 844 & 346 & 125 & 124 & -  \\
         O' Brien & 6,306 & 657 & 548 & 196 & 187 & -  \\
         Oliver &  10,330 & 739 & 670 & 118 & 98 & -  \\
    \end{tabular}
    }
    \vspace{0.2cm}
    \caption{Columns 2-3: Training data statistics in terms of the total number of unique words and the number of word-specific models that we train. Columns 4-7: number of unique word models tested in each sub-task.
    }
    \label{tab:testing}
    \vspace{-0.6cm}
\end{table}

We evaluate our approach on five falsification scenarios and compare it with the state-of-the-art deepfake detection methods and a phoneme-based baseline. \new{We also provide several ablations and analysis studies. We end with showcasing our method's interpretability.} %

\subsection{Implementation Details}

\textbf{Data Preprocessing:} Each video is first preprocessed so that only the person of interest is retained. Given one frame of an input video, we first use a single-stage face detector~\cite{Deng_2020_CVPR} to localize all the faces. Then a face recognition network ArcFace \cite{Deng_2019_CVPR} is used to check whether each face is the target person, and the outliers are masked out. \new{(For impersonator videos, the face of impersonator is used instead of the target person.) }%
For transcription, we used an open-source implementation of DeepSpeech~\cite{hannun2014deep}. For AU extraction, we use the facial behavior analysis toolkit OpenFace2~\cite{baltruvsaitis2015cross,baltrusaitis2018openface}. %

\begin{table}[t]
    \centering
    \resizebox{0.48\textwidth}{!}
	{
    \begin{tabular}{l|c|c|c|c|c}
    
         & Dubbing & Wav2Lip & Impersonator & FaceSwap & itw \\
         \hline
         Obama & 1.00& 1.00& 0.95& 0.90& 0.98 \\
         Trump & 0.95& 0.99& 0.89& 0.92& 0.98 \\
         Biden & 0.84& 0.93& 0.98& 0.73& 0.95 \\
         Harris & 0.90& 0.89& 0.82& 0.93 & -\\
         O' Brien & 0.91& 0.88 & 0.90& 0.84 & - \\
         Oliver & 0.94& 0.93 & 0.86& 0.87 & - \\
         \hline
         Avg & 0.92 & 0.94 & 0.90 & 0.87 & 0.97 \\
    \end{tabular}
    }
    \vspace{0.2cm}
    \caption{Accuracy in terms of AUC on 10-second video clips for the six individuals and five video falsification scenarios. The average AUC across individuals is given in the last row. }
    \label{tab:results}
    \vspace{-0.6cm}
\end{table}

\noindent\textbf{Training Details:} In our experiments, we use logistic regression to solve the binary classification problem of real/fake video. To train our person-specific word classifiers, we use 90\% of the real videos for the ``Real'' class, and 90\% of the Dubbing and Wav2Lip lip-sync videos for the ``Fake'' class. The number of unique words present in the speech for each individual is given in Table~\ref{tab:testing} (column 2). The word-specific models are trained for the words with average frequency of once per hour in training dataset. \new{For example, the overall duration of videos for Harris/Oliver is 7.5/53.4 hours. Therefore, in the case of Harris/Oliver, a word classifier is trained if the word frequency is greater than equal to 7/53, respectively.}
Shown in Table~\ref{tab:testing} (column 3) is the total number of word models trained for each individual. On average, 799 word models are trained, with the smallest/largest number of models trained for O'Brien/Obama.

\noindent\textbf{Testing Details:} We test our approach on remaining 10\% of real, audio dubbing and Wav2Lip lip-sync videos. Additionally, we test on all the videos with Impersonators, FaceSwap, and in-the-wild lip-sync deepfakes which were \emph{not seen during training} (as introduced in Section~\ref{sec:dataset}). Each test video is divided into overlapping 10-second video clips (30 fps) with a shift window of two seconds. 
Shown in the Table~\ref{tab:testing} (columns 4-7), is the total number of unique words that were evaluated in each of the test datasets, based on the occurred words within the trained words-set in testing time.

\noindent\textbf{Evaluation Metric:} We report the Area Under the Curve (AUC) score for the 10-second test videos. For the previous methods that perform analysis on a  temporal window less than 10 seconds, we average predictions over 10 seconds. 

\noindent\textbf{Methods:} 
As our approach does not analyze the audio signal for our detection model and uses only person-specific visual features conditioned on words, we compare our approach with other visual feature-based forensic techniques.  \new{Moreover, while there is prior work on audio-visual deepfake detection, we were unable to find any publicly available code bases.}  Thus, we select the following prior methods with available code bases: the low-level feature-based method in XceptionNet~\cite{rossler2019faceforensics++}; the high-level semantic-based approach in LipForensics~\cite{haliassos2021lips}; the biometric-based techniques in Protecting World Leaders (PWL)~\cite{agarwal2019} and ID-Reveal~\cite{cozzolino2020id}. 
\new{At the same time, we are interested in empirically assessing whether using words provides some additional benefits over using sounds in the audio. To address that, we construct a version of our method that uses \emph{phonemes}\footnote{We use the CMU Pronouncing Dictionary (https://github.com/cmusphinx/cmudict) which breaks the words from video transcripts into phonemes, of which there are $70$. } instead of words for determining visual windows and training classifiers. Since phonemes correspond to sounds made during speech, this serves as a proxy to audio-visual methods.} 

\subsection{Results}
\label{sec:results}

Shown in Table~\ref{tab:results}, is the performance of our method in terms of AUC for each individual test case. The average AUC across all the individuals is shown in the bottom row. Our approach works the best for Obama with the average AUC of 0.97 across all types of falsification scenarios and the worst for O'Brien with an average AUC of 0.88. This is expected as the Obama videos have higher quality and better consistency in facial movement during the formal weekly addresses. The videos of O'Brien are of lower visual quality and have a wider range of facial movements during the informal interviews, monologues, and audience interactions during the talk-show. This makes it more difficult for our word-conditioned model to learn consistent facial movement patterns from O'Brien videos. 

\noindent\textbf{Comparison with State-Of-The-Art:} 
Shown in Table~\ref{tab:related} are the average AUCs across all the individuals for each method and video falsification scenarios. Our approach performs the best across all the video falsification scenarios except in case of Wav2Lip where LipForensics obtains the best performance of 0.98. All the previous methods fail to detect the dubbing video falsification scenario as there is no video manipulation performed in this case. The non-biometric techniques fail to detect impersonators' video. Even though the related  biometrics-based methods are able to detect FaceSwaps and impersonators, they perform poorly on lip-sync videos. This is because these techniques only use the visual cues of a person identity, most of which are preserved in the lip-sync videos. This shows the advantage of our approach, i.e. using words in combination with the visual cues. \new{When comparing word- to phoneme-conditioning, we see that phonemes have strong ability to detect audio-visual inconsistency, but fail in capturing person-specific features needed for Impersonator and FaceSwap fakes. This is intuitive, since phonemes correspond to sounds made over short spans and shared across many words, thus missing some semantic and idiosyncratic clues that can be leveraged via word-conditioning.
To sum up, word-conditioning enables us to capture both the audio-visual inconsistency and the biometric features.}

\begin{table}[t]
    \centering
    \resizebox{0.48\textwidth}{!}
	{
    \begin{tabular}{l|c|c|c|c|c}
         & Dubbing & Wav2Lip & Impersonator & FaceSwap & itw \\
         \hline
         XceptionNet~\cite{rossler2019faceforensics++} &    0.50 &   0.78 & 0.57 &   0.54 & 0.49 \\
         LipForensics~\cite{haliassos2021lips} &    0.50 &    \textbf{0.98}&     0.43 &  0.81 &     0.95 \\
         PWL~\cite{agarwal2019} &    0.50&     0.63&     0.86&     0.85&     0.60 \\
         ID-Reveal~\cite{cozzolino2020id} &   0.50 &   0.66 &     0.85      & 0.78 & 0.61 \\
         \hline
         \new{Ours w/ Phonemes} & \new{\textbf{0.95}} & \new{{0.96}} & \new{0.61}& \new{0.58}& \new{\textbf{0.98}} \\
         Ours w/ Words & {0.92} & 0.94 & \textbf{0.90} & \textbf{0.87} & {0.97} \\
    \end{tabular}
    }
    \vspace{0.2cm}
    \caption{The performance in terms of AUCs on 10-second video clips. For each method and video falsification scenario, shown above are the average AUCs across all six individuals.}
    \label{tab:related}
\end{table}
\begin{table}[t]
    \centering
    \resizebox{0.48\textwidth}{!}
	{
    \begin{tabular}{p{25mm}|c|c|c|c|c}
    
         & Dubbing & Wav2Lip & Impersonator & FaceSwap & itw \\
         \hline
         Fixed Window & 0.50& 0.91& 0.81& 0.68& 0.87 \\
         Word Window & 0.79& 0.88& 0.72& 0.68& 0.94 \\
         \hline
         Ours w/ Words & 0.92 & 0.94 & {0.90} & {0.87} & 0.97 \\
    \end{tabular}
    }
    \vspace{0.2cm}
    \caption{The average AUC performance across all individuals for two ablations of our method, see text for details.
    }
    \label{tab:words_use}
\end{table}

\begin{figure}[t]
    \begin{center}
    \begin{tabular}{cc}
          \includegraphics[width=0.45\linewidth]{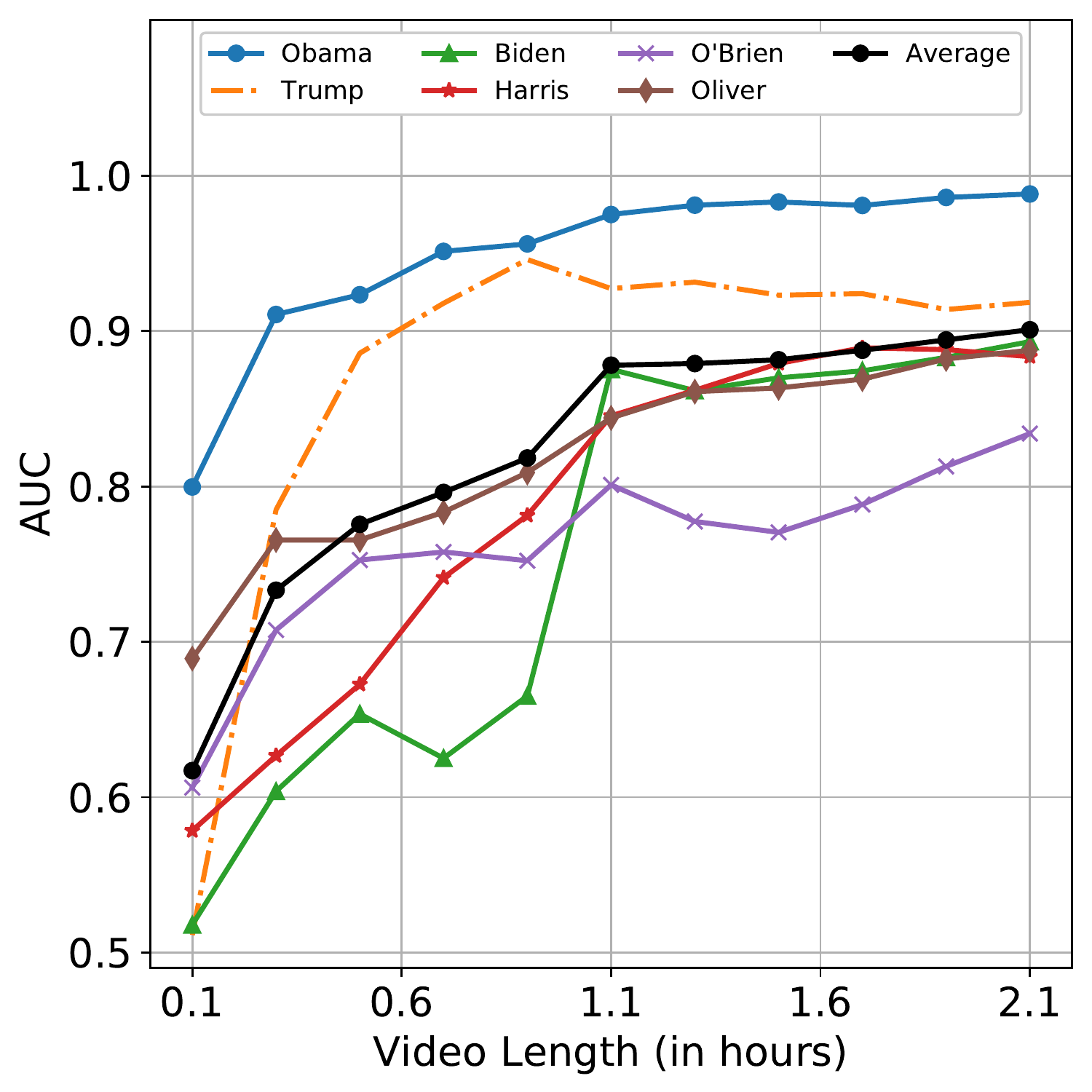} &
        \includegraphics[width=0.45\linewidth]{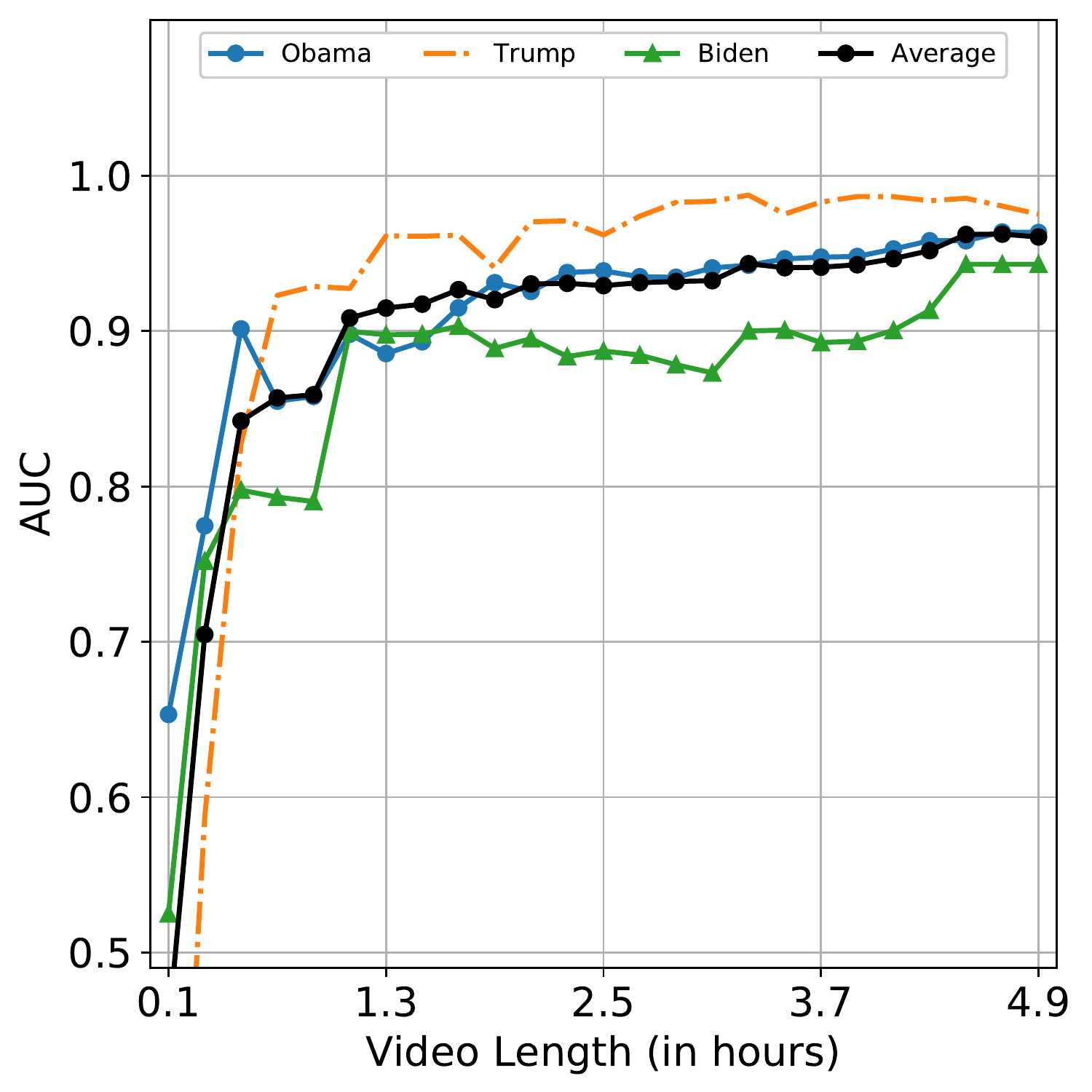} \\
    \end{tabular}
    \end{center}
    \vspace{-0.3cm}
    \caption{Effect of number of hours of training video. Left plot is evaluated on Wav2Lip fakes, the right one is evaluated on the in-the-wild fakes.}
\label{fig:time_ablation}
\end{figure}

\begin{table}[t]
    \centering
    \resizebox{0.48\textwidth}{!}
	{
    \begin{tabular}{l|c|c|c|c|c|c}
    
         Wav2Lip & Obama & Trump & Biden & Harris & O'Brien & Oliver \\
         \hline
         \new{Ours w/ Phonemes} & \new{0.73} & \new{0.66} & \new{0.63} & \new{0.67} & \new{0.69} & \new{0.81} \\
         \new{Ours w/ Words} & \new{0.68}	& \new{0.65} & \new{0.62} & \new{0.64} & \new{0.56} & \new{0.72} \\ 
    \end{tabular}
    }
    \vspace{0.2cm}
    \caption{\new{``Transfer'' performance in terms of AUCs for models trained on five people and tested on a held-out sixth person (pristine vs. Wav2Lip), reported per test person, averaged.}} 
    \label{tab:transfer}
    \vspace{-0.6cm}
\end{table}

\begin{figure*}[ht]
    \begin{center}
    \resizebox{0.9\textwidth}{!}
	{
        \begin{tabular}{l@{\hspace{0.1cm}}c@{\hspace{0.1cm}}c@{\hspace{0.1cm}}c@{\hspace{0.1cm}}c|c@{\hspace{0.1cm}}c@{\hspace{0.1cm}}c@{\hspace{0.1cm}}c@{\hspace{0.1cm}}c}
        & \multicolumn{8}{c}{chin-raise (AU17) and lip-rounding (lip-hor) during the word ``tremendous''} &\\
            \rotatebox[origin=lB]{90}{real}
            &\includegraphics[width=0.1\linewidth]{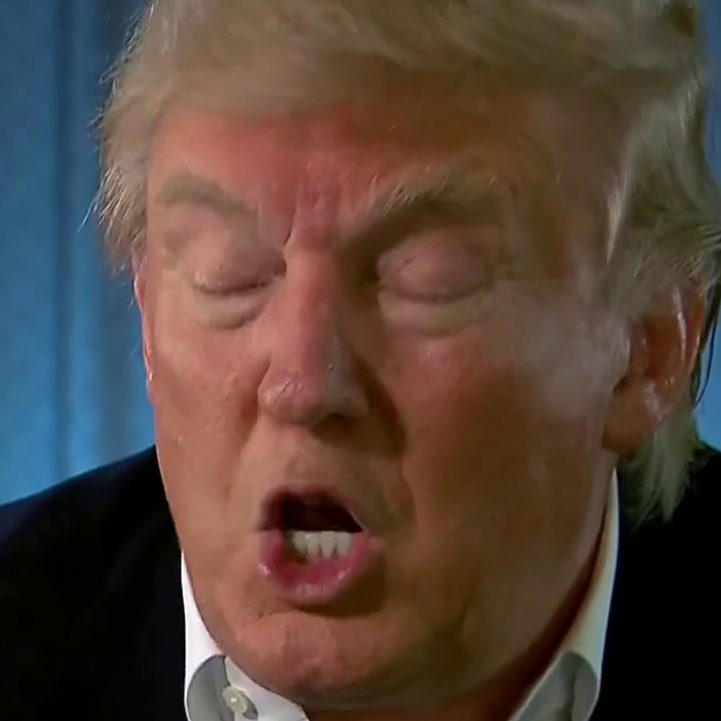} 
           & \includegraphics[width=0.1\linewidth]{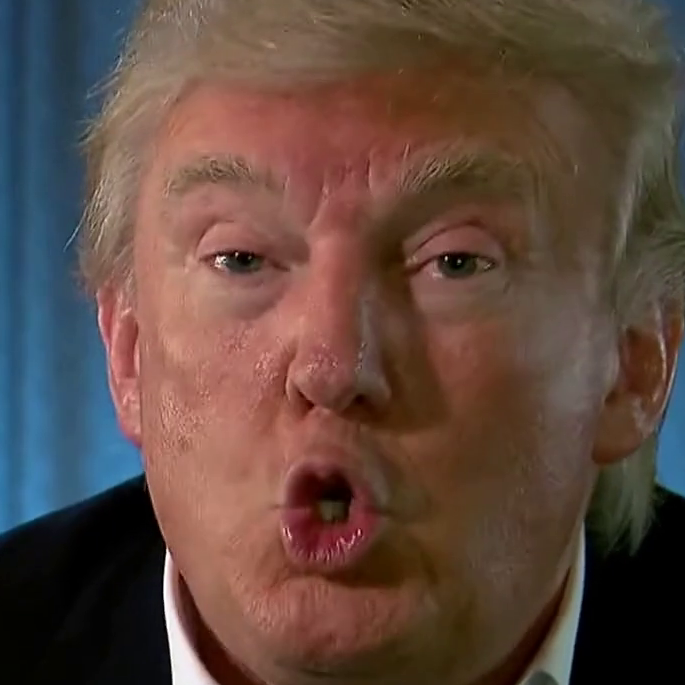} 
           & \includegraphics[width=0.1\linewidth]{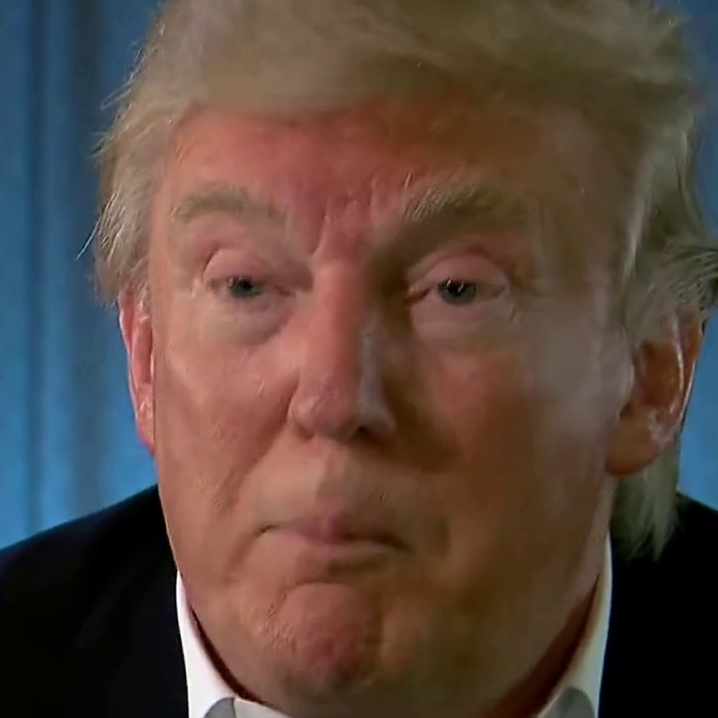}
           & \includegraphics[width=0.1\linewidth]{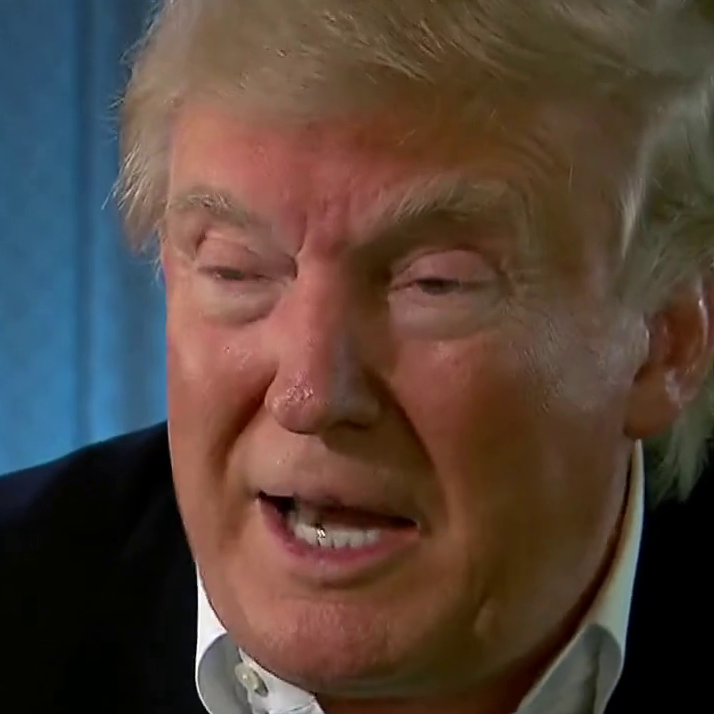}
           & \includegraphics[width=0.1\linewidth]{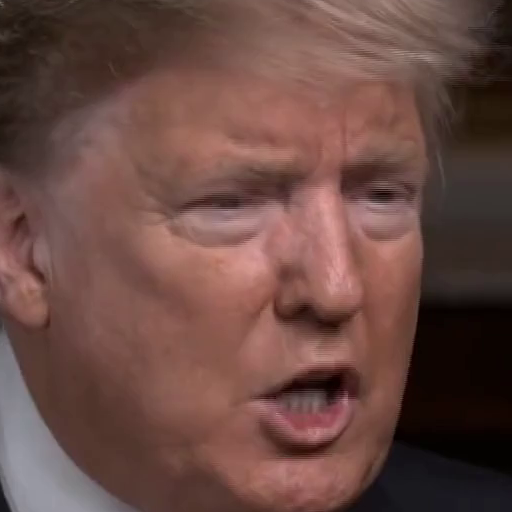} 
            & \includegraphics[width=0.1\linewidth]{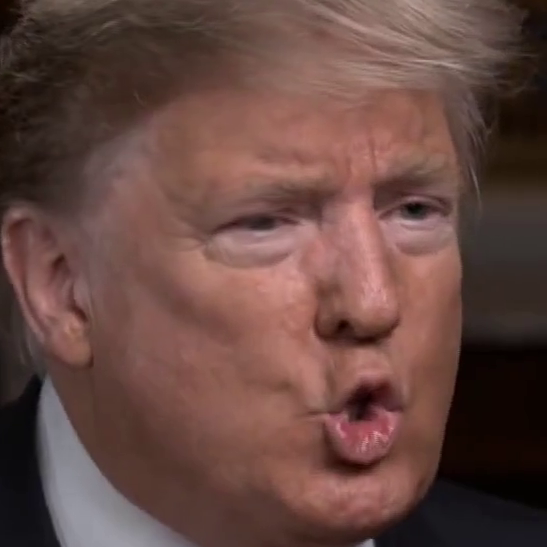} 
            & \includegraphics[width=0.1\linewidth]{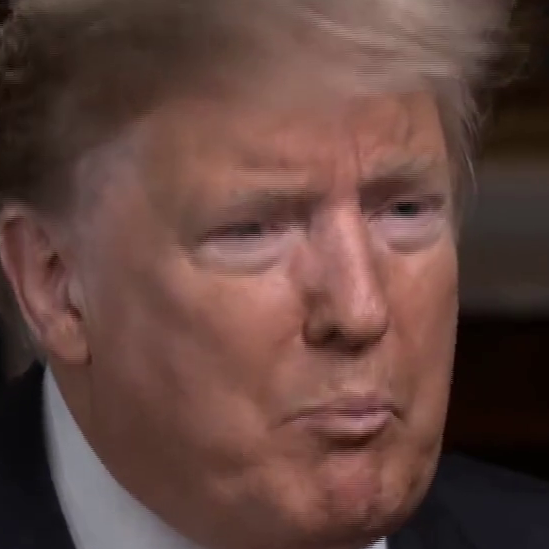}
            & \includegraphics[width=0.1\linewidth]{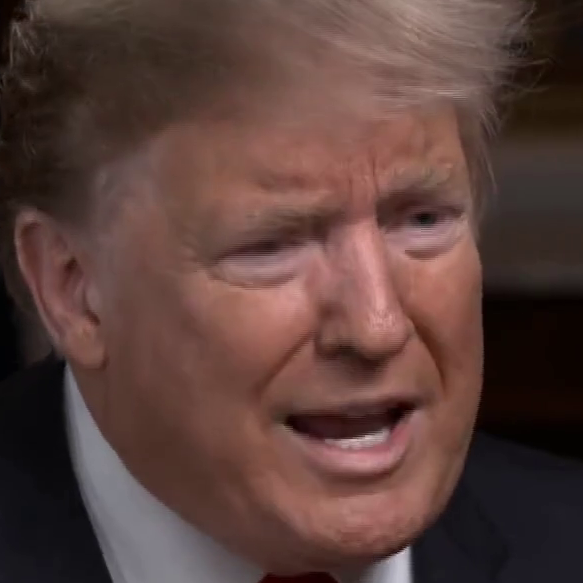} 
            &\includegraphics[width=0.1\linewidth]{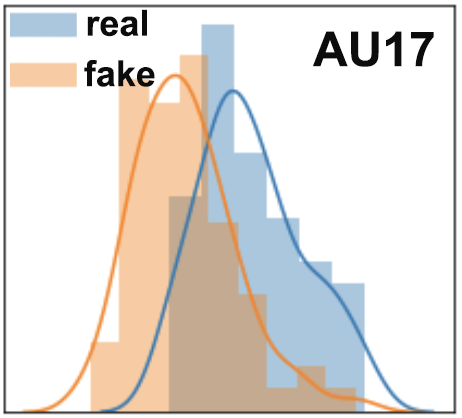}\\
           
           \rotatebox[origin=lB]{90}{fake}
           & \includegraphics[width=0.1\linewidth]{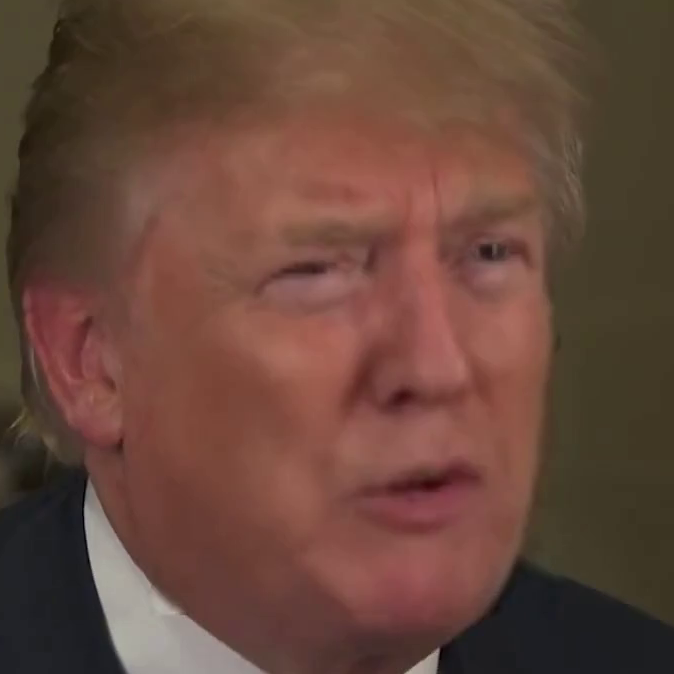} 
           & \includegraphics[width=0.1\linewidth]{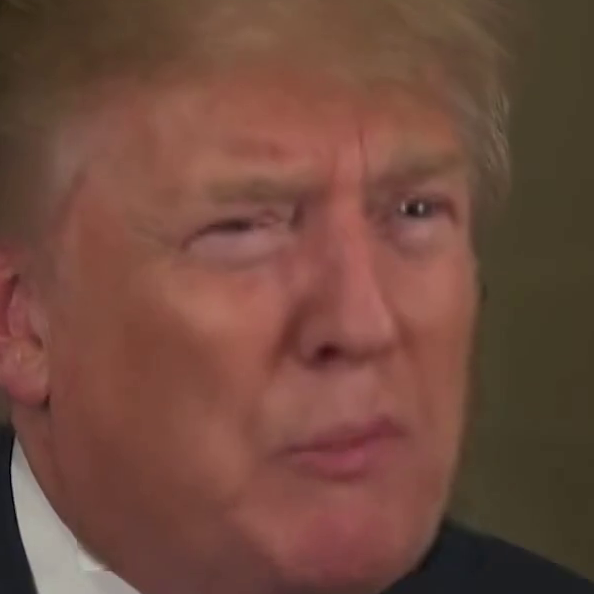} 
           & \includegraphics[width=0.1\linewidth]{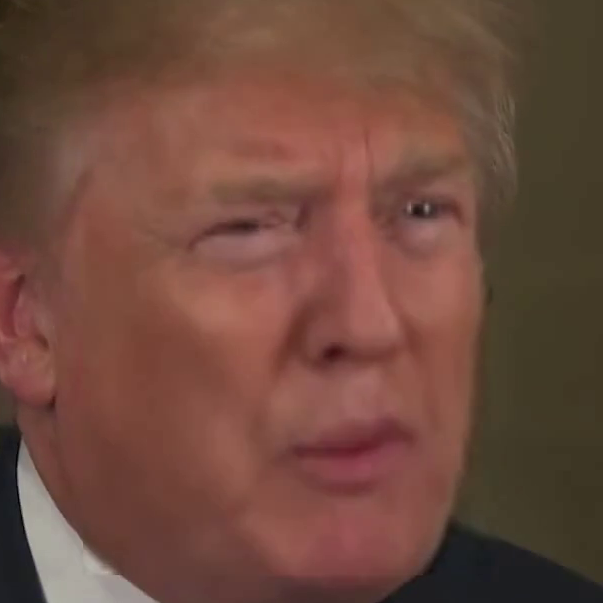}
           & \includegraphics[width=0.1\linewidth]{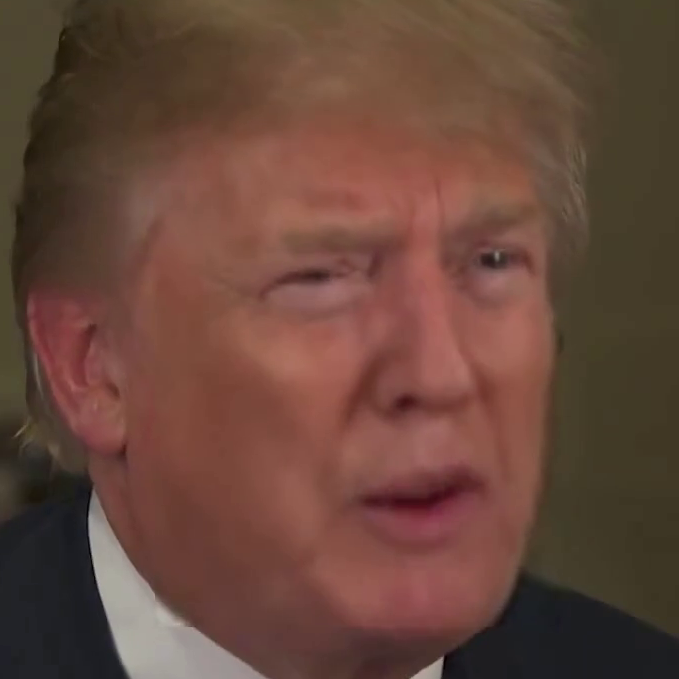}
            & \includegraphics[width=0.1\linewidth]{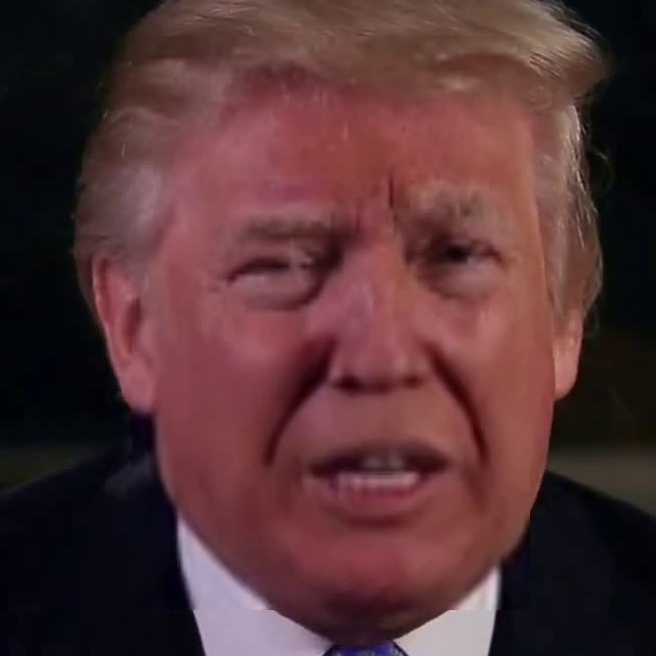} 
            & \includegraphics[width=0.1\linewidth]{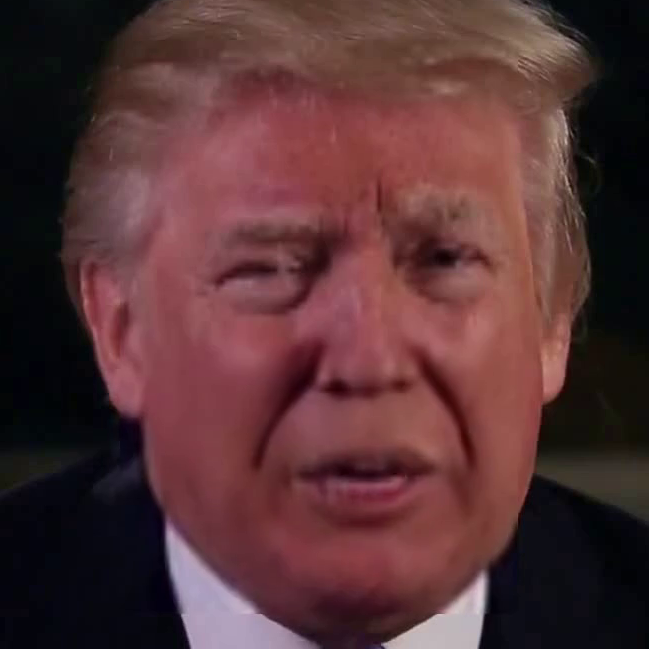} 
            & \includegraphics[width=0.1\linewidth]{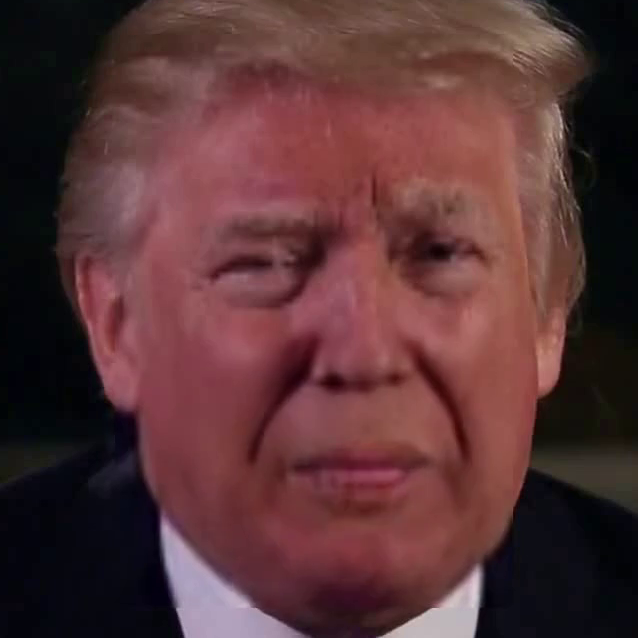}
            & \includegraphics[width=0.1\linewidth]{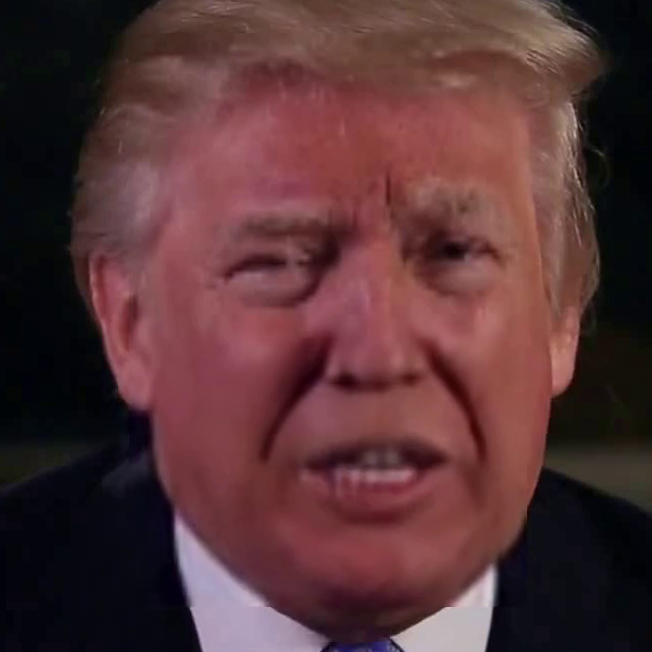}
            &\includegraphics[width=0.1\linewidth]{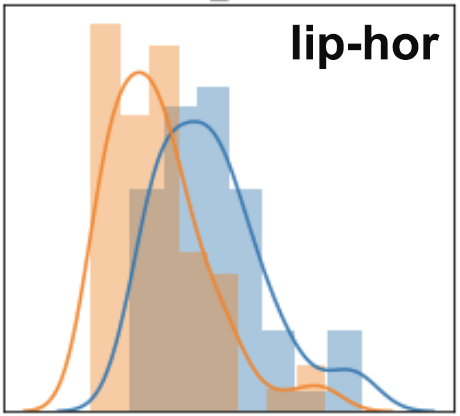}\\
         & \multicolumn{8}{c}{dimpler (AU14) and lip-corner-pull (AU12) during the word ``billion''} & \\
         \rotatebox[origin=lB]{90}{real}
            &\includegraphics[width=0.1\linewidth]{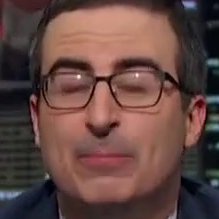} 
           & \includegraphics[width=0.1\linewidth]{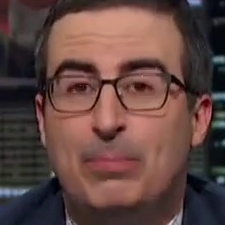} 
           & \includegraphics[width=0.1\linewidth]{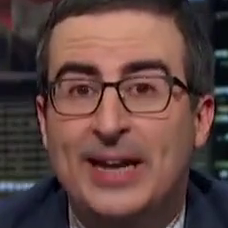}
           & \includegraphics[width=0.1\linewidth]{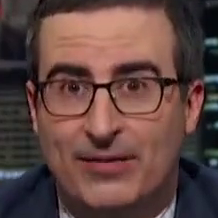}
           & \includegraphics[width=0.1\linewidth]{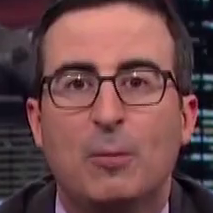} 
            & \includegraphics[width=0.1\linewidth]{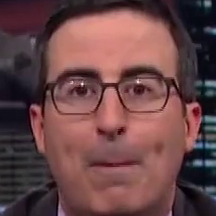} 
            & \includegraphics[width=0.1\linewidth]{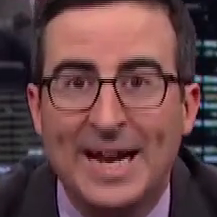}
            & \includegraphics[width=0.1\linewidth]{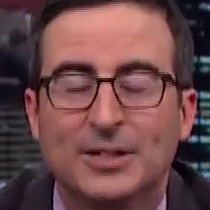}
           &\includegraphics[width=0.1\linewidth]{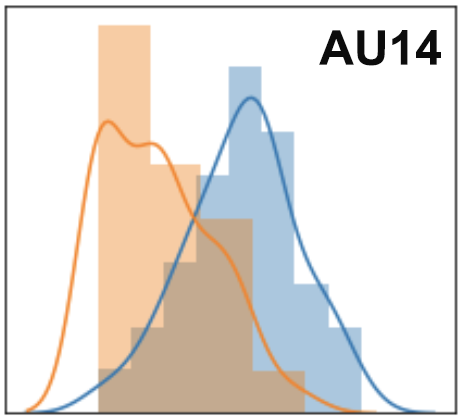}\\

            \rotatebox[origin=lB]{90}{fake}
           &  \includegraphics[width=0.1\linewidth]{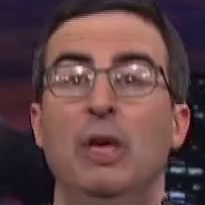} 
           & \includegraphics[width=0.1\linewidth]{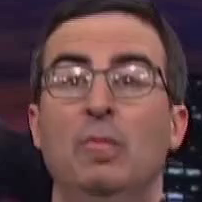} 
           & \includegraphics[width=0.1\linewidth]{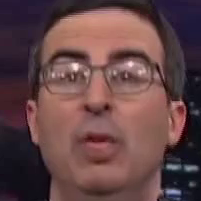}
           & \includegraphics[width=0.1\linewidth]{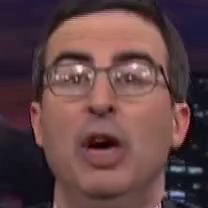}
            &  \includegraphics[width=0.1\linewidth]{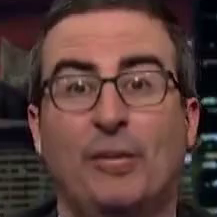} 
           & \includegraphics[width=0.1\linewidth]{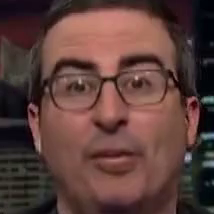} 
           & \includegraphics[width=0.1\linewidth]{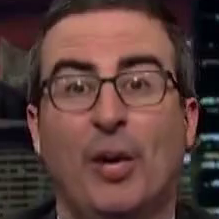}
           & \includegraphics[width=0.1\linewidth]{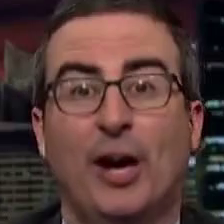}
           &\includegraphics[width=0.1\linewidth]{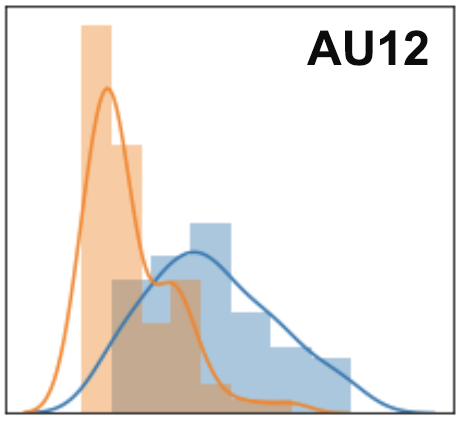}\\
        \end{tabular}
        }
    \end{center}
    \vspace{-0.3cm}
\caption{Qualitative examples of the facial movement for specific words that were used to predict real vs. fake. For each word and individual, we show two examples of facial movement from real and fake (Wav2Lip) videos. In the last column we show the distribution of a gesture feature in real and fake training dataset of the individual. E.g., for Trump, the lip rounding and chin raise actions during the word ``tremendous'' are missing in the fakes. This is supported by the distributions of AU17 and lip-hor AU: the average strength of these movements is lower in fake videos of Trump than the real ones.}
\label{fig:qualitative}
\vspace{-0.6cm}
\end{figure*}

\noindent\textbf{Effect of Using Words:} We further analyse the effect of training the word-specific classifiers by training two different versions of our approach. In the first version (Fixed Window), we do not use the word information and compute the 25-D visual gesture features using all the non-overlapping fixed windows of 30 frames. This window size is chosen as 95\% of the words have a duration smaller or equal to 30 frames. Using the corresponding gesture features, we train a single linear classifier to predict Real vs. Fake. In the second version (Word Window), the gesture features are extracted using word intervals, as in our approach, but we train a single linear classifier instead of word-specific classifiers. 
Shown in Table~\ref{tab:words_use}, are the average AUCs across all individuals for the two ablations and our approach. While the word intervals already improve over the fixed window case, the word-specific training helps improve the performance on each type of video falsifications. This clearly shows that the key advantage of our approach is indeed in leveraging the word-conditioned facial gesture analysis. 

\noindent\textbf{Model Transfer and Person Specific Features:} \new{Based on the previous experiments, we can already hypothesize that our word-conditioned method captures some person-specific features, as evident from its high performance on Impersonator and FaceSwap falsified videos. To further analyze the degree of ``person-specificity'', we conduct a model ``transfer'' experiment. Namely, we use models trained for five people and test them on a held-out sixth person. Intuitively, we expect these models not to do well in distinguishing the real vs. Wav2Lip fakes of that person. The average AUC scores are reported in Table~\ref{tab:transfer}. First, comparing the scores to Table~\ref{tab:results}, we see that the overall ``transfer'' is rather poor, as expected (e.g., a model trained for Obama achieves AUC 1.0 when tested on Obama, while the ``transferred'' models only give AUC 0.68.) Second, the phoneme-conditioning consistently gets higher ``transfer'' scores, showing that while it is person-specific, it captures more person-agnostic features, e.g., generic sound-to-lips alignment.}

\noindent\textbf{Effect of Training Data Size:}
We analyse the effect of number of hours of real videos used for training person-specific word models. The effect of training size is evaluated on: 1) Wav2Lip lip-sync fakes which on average have 72\% vocabulary overlap with the training dataset and 2) in-the-wild lip-sync fakes which on average have only 28\% vocabulary overlap with the training dataset.

Shown in Figure~\ref{fig:time_ablation} are the AUCs for individuals as the function of training size ranging from 0.1 to 2.1/5.0 hours of real training videos. For each real training size, we use the equal number of hours of fake videos from audio dubbing and Wav2Lip training datasets. The evaluation in the left/right plot is performed on Wav2Lip/in-the-wild lip-sync fakes. Shown with the black curve is the average AUC across all individuals as a function of training size. In each of these evaluation scenarios, the performance improves with the number of hours in training. In case of Wav2Lip, the average performance improves from 0.62 to 0.88 (42\%) from 0.1 to 1.3 hours and then improves from 0.88 to 0.90 (2.0\%) for training size greater than 1.3 hours. Similarly, for the in-the-wild lip-sync fakes, the average performance of 0.91 is achieved with 1.3 hours of training videos with only a slight improvement after that. This shows that while we used several hours of video per individuals, a relatively smaller training dataset ($\approx$1.5 hours) can provide a similar performance.

\noindent\textbf{Qualitative Results and Interpretability:} Here we present qualitative results showing the regularity of facial movements associated with words. Shown in Figure~\ref{fig:qualitative} are word-based facial movements of two individuals. %
For each one, we select one word from the top-5 performing words. (The performance of the word-based classifiers is evaluated on our training data in terms of word-level AUC.) For each selected word and individual, two occurrences are shown, from real (top row) and Wav2Lip fake (bottom row) videos. Shown in the last column is the distribution of one gesture feature (AU) in real and fake training data. 
We see that Trump, while saying the word ``tremendous'', rounds his lips and then presses the lips together before finally opening the lips apart. This rounding action of the lips is absent in the fake examples, even though the lips are closed once in the sequence. This difference in real and fake utterances can also be seen in the distributions of change in chin-raise (AU17) and change in ``lip-hor'' AUs. For Oliver, the word ``billion'' is associated with the creation of dimples on the cheeks, which is violated in the fake frames shown here. %
Thus, in addition to showing good generalization across a range of video falsifications, our method provides interpretability, offering insight into what words/gestures may be responsible for classifying a video as a fake. This is an important capability for an analyst using this tool.

%% file: conclusion.tex
\section{Discussion and Limitations}
\label{sec:conclusion}

We proposed a novel multi-modal, semantic-based approach for detecting falsified videos. We leverage the idea of learning person-specific associations between the speaker's facial gestures and spoken words to verify the purported person's identity. 
Our experiments show that inconsistent head movements and facial expressions can be identified reliably when an impersonator is used for falsification. Moreover, we demonstrate the effectiveness and robustness of our approach on a wide range of deep and cheapfakes, outperforming all other methods in most cases. Since we do not attempt to detect video manipulation artifacts, our method will still work for more advanced future deepfakes. While other multi-modal detection techniques have shown that audio is an important cue for revealing falsifications, our semantic approach of using words can be an important addition, especially for cases when similar sounding words with different meanings are used. \new{Our experiments with word vs. phoneme conditioning support that.}

Our current approach relies on the accuracy of 3-D facial tracking via AU extraction.
While this is feasible for our dataset, where the speakers are often front-facing, for unconstrained videos, deep learning based features may be more reliable.
Although our method seems to behave rather sample efficiently (Figure~\ref{fig:time_ablation}), it is person-specific and thus requires sufficient ($\approx$ $1$ hour) training data to be effective. This data requirement can easily be satisfied for celebrities and world-leaders who are the most vulnerable to deep-fake attacks. 
Furthermore, while the AUs allow us to obtain interpretable results, denser 3-D facial features could allow for detecting more subtle anomalies. %
Finally, we have only validated our method for English speech. In the future, we would like to explore how well our word-conditioned technique would work with other languages.

Falsified media is a threat to society, so we envision positive impact from our work. At the same time, almost any method for fake detection may be adapted to create more robust fakes. As the visual quality of fakes keeps improving, it will be increasingly important to build such biometric models to mitigate the harm of deepfakes.

\noindent\textbf{\textit{Acknowledgements.}} This work was supported in part by DoD including DARPA’s SemaFor, PTG and/or LwLL programs, as well as BAIR’s industrial alliance programs. We thank Sarthak Kamat for his valuable feedback during the project.

%% file: appendix_arxiv.tex
\section*{\label{sec:appendix} Supplementary Material}

We first evaluate the robustness of our system against unseen video perturbations (Section~\ref{sec:supp_robustness}). Next, we provide some quantitative and qualitative results to support the analysis made in the main paper. In the main paper we compared with the related works using the average AUCs across all individuals. In order to give a better insight into the comparison, we first present per-individual results for each of the related works (Section~\ref{sec:supp_sota}). We then present a more detailed version of qualitative results using both training and testing datasets (Sections~\ref{sec:supp_video} and \ref{sec:supp_words}). 

\section{Robustness Test}
\label{sec:supp_robustness}
For this experiment, we re-saved the real test videos of each individual using the ffmpeg compression quality of 40. Shown in Table~\ref{tab:compression} are the results when compared to the best performing related techniques. In our case the average performance is reduced by 5\% from 0.92 to 0.87, whereas in case of LipForensics and PWL the reduction is 14\% and 6\%. Even though our performance is reduced, our approach still performs better than previous techniques. 

\begin{table}[h]
    \centering
    \resizebox{0.48\textwidth}{!}
	{
    \begin{tabular}{l|c|c|c|c|c}
         & Dubbing & Wav2Lip & Impersonator & FaceSwap & itw \\
         \hline
         LipForensic &    0.50 & 0.85 & 0.34 & 0.67 & 0.79 \\
         PWL &    0.50&     0.58&     \textbf{0.86}&     \textbf{0.79}&     0.53 \\
         \hline
         Ours & \textbf{0.89}&\textbf{ 0.91}& 0.84& \textbf{0.79}& \textbf{0.94} \\
    \end{tabular}
    }
    \caption{The performance in terms of AUCs on 10-second video clips after compression of real test videos. }
    \label{tab:compression}
    \vspace{-0.5cm}
\end{table}

\section{Comparison with State-Of-The-Art}
\label{sec:supp_sota}

\begin{table}[t]
        \centering
        \resizebox{0.48\textwidth}{!}
    	{
        \begin{tabular}{l|c|c|c|c|c}
            \hline
            \multicolumn{6}{c}{XceptionNet} \\
            \hline
             & Audio &  &  &  &  \\
             & Dubbing & Wav2Lip & Impersonator & FaceSwap & in-the-wild \\
             \hline
             Obama & 0.50& 0.94& 0.74& 0.96& 0.47 \\ 
             Trump & 0.50& 0.84& 0.70& 0.82& 0.54 \\ 
             Biden & 0.50& 0.49& 0.69& 0.67& 0.45 \\ 
             Harris & 0.50& 0.80& 0.48& 0.24 & -\\ 
             O' Brien & 0.50& 0.69& 0.44& 0.11 & - \\ 
             Oliver & 0.50& 0.93& 0.26& 0.15 & - \\ 
             
            \hline
            \multicolumn{6}{c}{PWL} \\
            \hline
             & Audio &  &  &  &  \\
             & Dubbing & Wav2Lip & Impersonator & FaceSwap & in-the-wild \\
             \hline
             Obama & 0.5& 0.56& 0.96& 0.96& 0.83 \\ 
             Trump & 0.5& 0.51& 0.95& 0.94& 0.41 \\ 
             Biden & 0.5& 0.53& 0.65& 0.66& 0.55 \\ 
             Harris & 0.5& 0.45& 0.94& 0.94&  -\\ 
             O' Brien & 0.5& 0.84& 0.69& 0.67& - \\ 
             Oliver & 0.5& 0.88& 0.99& 0.93& - \\ 
        
            \hline
            \multicolumn{6}{c}{LipForensics} \\
            \hline
             & Audio &  &  &  &  \\
             & Dubbing & Wav2Lip & Impersonator & FaceSwap & in-the-wild \\
             \hline
             Obama & 0.50& 1.00& 0.83& 1.00& 0.98 \\ 
             Trump & 0.50& 1.00& 0.68& 0.98& 0.97 \\ 
             Biden & 0.50& 0.93& 0.15& 0.30& 0.91 \\ 
             Harris & 0.50& 1.00& 0.08& 0.71 & -\\ 
             O' Brien & 0.50& 0.96& 0.48& 0.90 & - \\ 
             Oliver & 0.50& 0.97& 0.39& 0.98 & - \\ 
             
            \hline
            \multicolumn{6}{c}{ID-Reveal} \\
            \hline
             & Audio &  &  &  &  \\
             & Dubbing & Wav2Lip & Impersonator & FaceSwap & in-the-wild \\
             \hline
             Obama & 0.50& 0.77& 0.81& 0.71& 0.59 \\ 
             Trump & 0.50& 0.66& 0.92& 0.88& 0.77 \\ 
             Biden & 0.50& 0.47& 0.75& 0.59& 0.47 \\ 
             Harris & 0.50& 0.73& 0.98& 0.98 & -\\ 
             O' Brien & 0.50& 0.66& 0.63& 0.56 & - \\ 
             Oliver & 0.50& 0.69& 0.98& 0.93 & - \\ 
             
        \end{tabular}
        }
        \caption{Accuracy in terms of AUC on 10-second video clips for the six individuals and five different video falsification scenarios. The average AUC across all individuals in given in the last row. From top-bottom are the AUCs for XceptionNet, PWL, LipForensics and ID-Reveal. }
        \label{tab:supp_sota}
    \end{table}

Shown in Table~\ref{tab:supp_sota} are the per-individual results for all the related methods that were presented in the main paper.

\section{Videos for Qualitative Analysis}
\label{sec:supp_video}

Here we provide the videos used for qualitative analysis of the words presented in the Figure 1 and Figure 6 of the main paper. For Obama, Trump, and Oliver we provide occurrences of the word ``hi'', ``tremendous'', and ``billion'' in the real and fake videos. Therefore, there are a total of six videos for this section:
\begin{itemize}
    \item \href{https://github.com/agarwalShruti15/wtw\_project\_page/blob/7aa9adbe3e33a60c1819481990e030da731f3d42/videos/Obama\_hi\_real.mp4}{Obama\_hi\_real.mp4},
    \item \href{https://github.com/agarwalShruti15/wtw\_project\_page/blob/7aa9adbe3e33a60c1819481990e030da731f3d42/videos/Obama\_hi\_fake.mp4}{Obama\_hi\_fake.mp4},
    \item  \href{https://github.com/agarwalShruti15/wtw_project_page/blob/7aa9adbe3e33a60c1819481990e030da731f3d42/videos/Trump_tremendous_real.mp4}{Trump\_tremendous\_real.mp4},
    \item \href{https://github.com/agarwalShruti15/wtw_project_page/blob/7aa9adbe3e33a60c1819481990e030da731f3d42/videos/Trump_tremendous_fake.mp4}{Trump\_tremendous\_fake.mp4},
    \item \href{https://github.com/agarwalShruti15/wtw_project_page/blob/7aa9adbe3e33a60c1819481990e030da731f3d42/videos/Oliver_billion_real.mp4}{Oliver\_billion\_real.mp4}, and 
    \item \href{https://github.com/agarwalShruti15/wtw_project_page/blob/7aa9adbe3e33a60c1819481990e030da731f3d42/videos/Oliver_billion_fake.mp4}{Oliver\_billion\_fake.mp4}.
\end{itemize}
In each video, the output probability of the word-specific classifier is shown in red on the top left corner (a value of 1 is for real and 0 is fake). The occurrences of the words are selected from the training dataset. This is done to demonstrate the facial gestures associated with specific words during training.

In each case, it can be observed that a specific facial gesture is present in real videos which is missing in the fake videos. For example, the occurrences of the word ``hi'' is associated with an upward head movement which is missing in the fake examples. Similarly, in case of the word ``tremendous'', notice the presence of lip rounding and chin raise action in multiple occurrences of the word in real videos, whereas these actions are missing in the fake videos.

\section{Word Analysis for in-the-wild videos}
\label{sec:supp_words}

Here we show how the results of our method can be interpreted during the evaluation of a test video. For this we provide four example videos, a real and a fake video of Obama and Trump. The real videos are from test-split of real dataset and fake videos are from in-the-wild dataset. The videos are named as:
\begin{itemize}
    \item \href{https://github.com/agarwalShruti15/wtw_project_page/blob/7aa9adbe3e33a60c1819481990e030da731f3d42/videos/Obama_itw_test.mp4}{Obama\_itw\_test.mp4},
    \item \href{https://github.com/agarwalShruti15/wtw_project_page/blob/7aa9adbe3e33a60c1819481990e030da731f3d42/videos/Obama_real_test.mp4}{Obama\_real\_test.mp4},
    \item \href{https://github.com/agarwalShruti15/wtw_project_page/blob/7aa9adbe3e33a60c1819481990e030da731f3d42/videos/Trump_itw_test.mp4}{Trump\_itw\_test.mp4}, and
    \item \href{https://github.com/agarwalShruti15/wtw_project_page/blob/7aa9adbe3e33a60c1819481990e030da731f3d42/videos/Trump_real_test.mp4}{Trump\_real\_test.mp4}.
\end{itemize}

Given a test video of 10-second length, we show the output of word-specific classifier for each word. Shown on the x-axis of the plot is time and on the y-axis is the probability that the word occurrence is real. Shown in orange is the probability of the word in the test video and shown in the blue is the average real probability of the word in real dataset during training. The region in blue indicates the standard deviation of training probability. The gaps in the plot indicate that the word-specific classifier was missing.  The current time is indicated by the red dot on the plot and the current word is displayed on the top of the video. 

These word-level probabilities, can be used to isolate the words which obtain low probability of being real. For example, in \\ Obama\_itw\_test.mp4 many words have a low probability of being real with a minimum probability of zero for the word ``coverage''. Similarly in Trump\_itw\_test.mp4 video, the word ``protected'' has the zero probability of being real. Whereas in the videos Obama\_real\_test.mp4 and Trump\_real\_test.mp4, the real probability for each of the words is close to training real dataset (average of 0.8). 

\begin{figure*}
    \begin{center}
          \includegraphics[width=\linewidth]{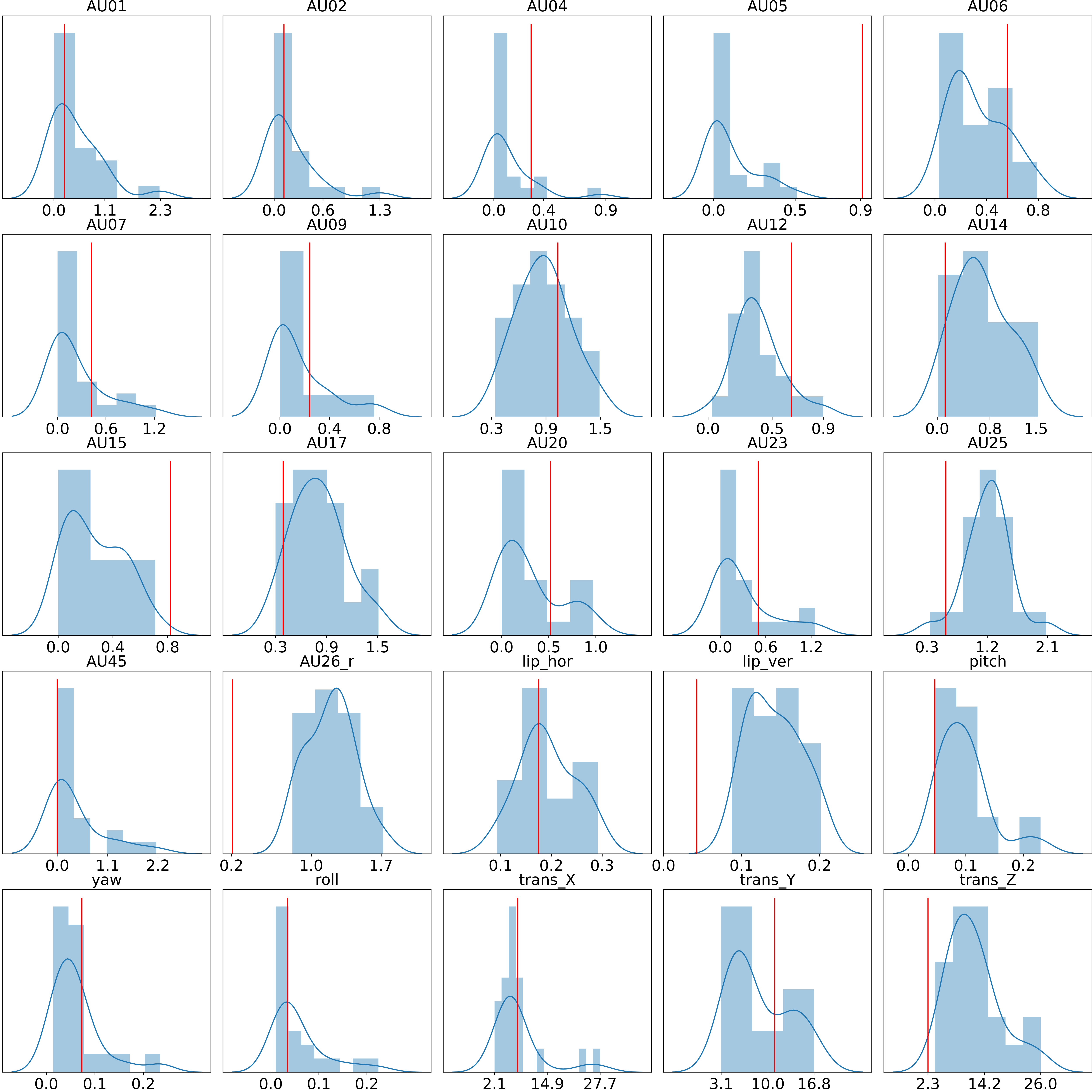}
    \end{center}
    \caption{In each panel, shown in blue is the distribution of one facial-gesture feature in real training videos of Obama for the word ``coverage''. The name of the facial-gesture feature is given on top of the panel. Shown with red line is the value of the facial feature in the current test video of Obama which in this case is the fake video shown in Obama\_itw\_test.mp4. }
    \label{fig:supp_obama_itw}
\end{figure*}

Shown in Figure~\ref{fig:supp_obama_itw} are the distributions of the 25 facial-gesture features for the word ``coverage'' for Obama. In each panel, shown in blue is the distribution of one facial-gesture feature in real training videos of Obama. Shown with red line is the value of facial-gesture feature in the current test video of Obama which in this case is the fake video shown in Obama\_itw\_test.mp4. The word ``coverage'' in this example fake video have an out-of-distribution value for AU26 i.e. jaw drop. The out-of-distribution value can also be observed for lip-ver motion where the value in the fake is lower than any of the value seen during training. 

\begin{figure*}
    \begin{center}
        \includegraphics[width=\linewidth]{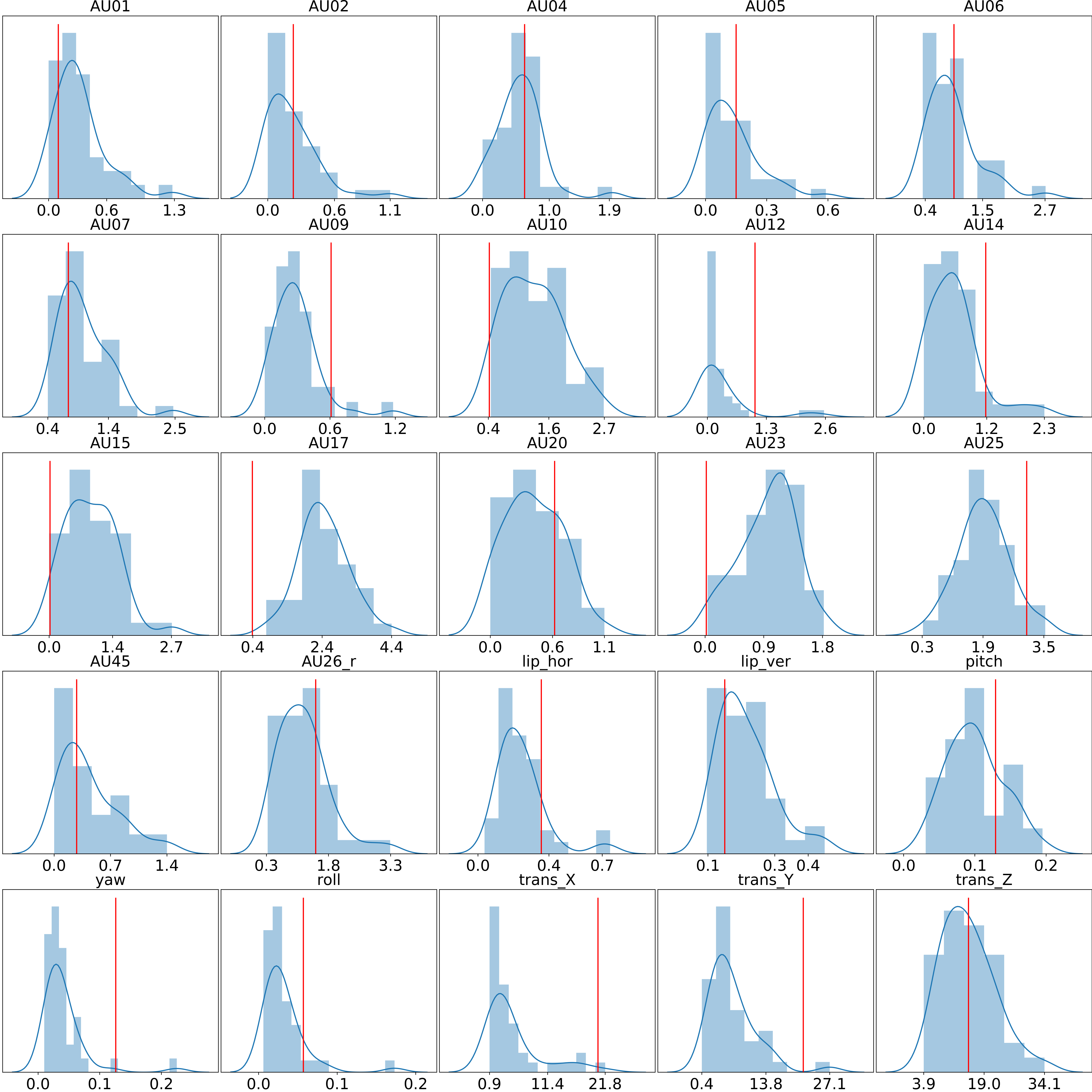}
    \end{center}
    \caption{In each panel, shown in blue is the distribution of one facial-gesture feature in real training videos of Trump for the word ``protect''. The name of the facial-gesture feature is given on top of the panel. Shown with red line is the value of the facial feature in the current test video of Trump which in this case is the fake video shown in Trump\_itw\_test.mp4. }
    \label{fig:supp_trump_itw}
\end{figure*}

Similarly, shown in Figure~\ref{fig:supp_trump_itw} are the distributions of the 25 facial-gesture features for the word ``protect'' for Trump. The red line in each panel is the value of facial-gesture feature in the fake test video of Trump shown in Trump\_itw\_test.mp4. For the word ``protect'' the value for AU17 (chin raise) and AU23 (lip tightner) in the fake is lower than any of the value seen during training.